\documentclass[sigconf]{acmart}
\AtBeginDocument{%
  \providecommand\BibTeX{{%
    \normalfont B\kern-0.5em{\scshape i\kern-0.25em b}\kern-0.8em\TeX}}}

\usepackage{multirow}
\usepackage{subcaption}
\usepackage{amsfonts,mathtools}

\usepackage{fixmath}

\copyrightyear{2022}
\acmYear{2022}
\setcopyright{acmcopyright}\acmConference[WSDM '22]{Proceedings of the Fifteenth ACM International Conference on Web Search and Data Mining}{February 21--25, 2022}{Tempe, AZ, USA}
\acmBooktitle{Proceedings of the Fifteenth ACM International Conference on Web Search and Data Mining (WSDM '22), February 21--25, 2022, Tempe, AZ, USA}
\acmPrice{15.00}
\acmDOI{10.1145/3488560.3498486}
\acmISBN{978-1-4503-9132-0/22/02}
\begin{document}
\fancyhead{}
\title{DAME: Domain Adaptation for Matching Entities
}

\author{Mohamed	Trabelsi}
\email{mot218@lehigh.edu}
\affiliation{%
 \institution{Lehigh University}
  \city{Bethlehem}
  \state{PA}
  \country{USA}
}

\author{Jeff Heflin}
\email{heflin@cse.lehigh.edu}
\affiliation{%
  \institution{Lehigh University}
  \city{Bethlehem}
  \state{PA}
  \country{USA}
}

\author{Jin Cao}
\email{jin.cao@nokia-bell-labs.com}
\affiliation{%
  \institution{Nokia Bell Labs}
  \city{Murray Hill}
  \state{NJ}
  \country{USA}
}

\begin{abstract}
Entity matching (EM) identifies data records that refer to the same real-world entity. Despite the effort in the past years to improve the performance in EM, the existing methods still require a huge amount of labeled data in each domain during the training phase. These methods treat each domain individually, and capture the specific signals for each dataset in EM, and this leads to overfitting on just one dataset. The knowledge that is learned from one dataset is not utilized to better understand the EM task in order to make predictions on the unseen datasets with fewer labeled samples. In this paper, we propose a new domain adaptation-based method that transfers the task knowledge from multiple source domains to a target domain. Our method presents a new setting for EM where the objective is to capture the task-specific knowledge from pretraining our model using multiple source domains, then testing our model on a target domain. We study the zero-shot learning case on the target domain, and demonstrate that our method learns the EM task and transfers knowledge to the target domain. We extensively study fine-tuning our model on the target dataset from multiple domains, and demonstrate that our model generalizes better than state-of-the-art methods in EM.
  
\end{abstract}




\maketitle

\section{Introduction}

\begin{figure*}
\centering
\begin{subfigure}{.5\textwidth}
  \centering
  \includegraphics[width=1\linewidth]{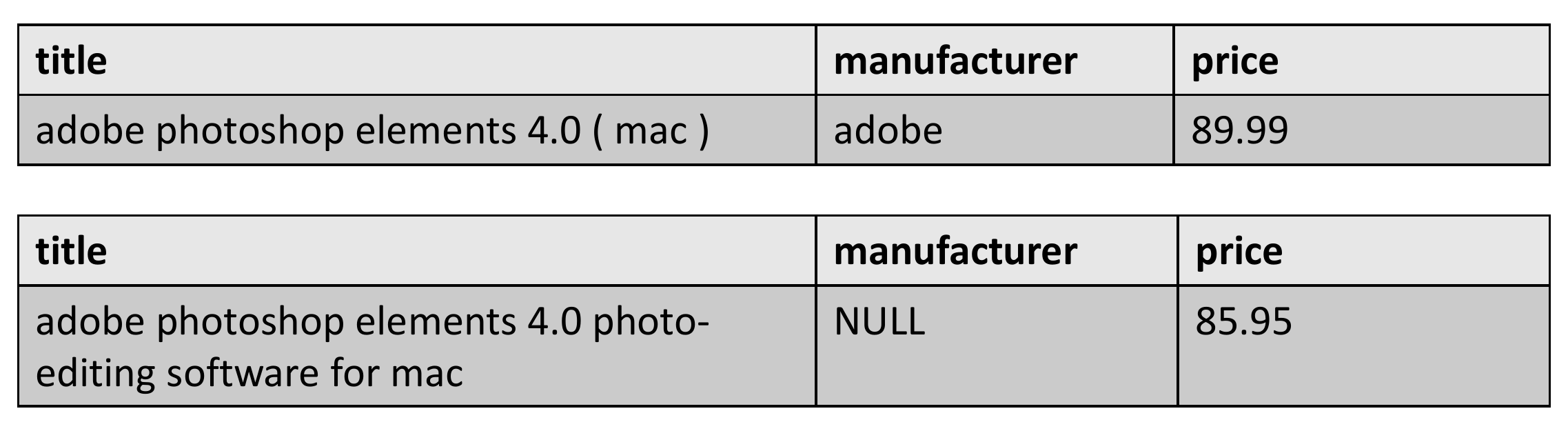}
  \caption{Records refer to the same entity: adobe photoshop 4.0}
  \label{fig:sub1}
\end{subfigure}%
\begin{subfigure}{.5\textwidth}
  \centering
  \includegraphics[width=1\linewidth]{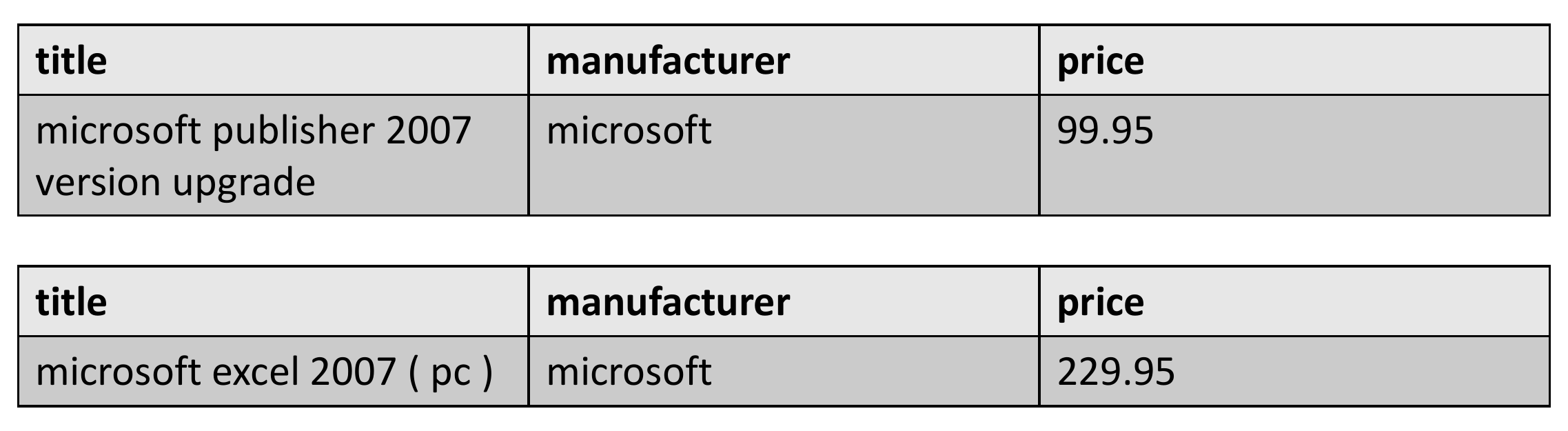}
  \caption{Records refer to different entities}
  \label{fig:sub2}
\end{subfigure}
\caption{Examples of pair of records for EM from Amazon-Google dataset. The
above record is from Amazon and the below record is from Google.}
\label{EM_examples}
\end{figure*}

Entity matching (EM) identifies data records that refer to the same real-world entity. EM is an important step in data cleaning and integration 
\cite{classifier1}, knowledge base enrichment \cite{kb_enrich}, 
and entity linking \cite{entity_linking}. Researchers have studied EM for many years in the context of data mining and integration.
In Figure \ref{EM_examples}, we show examples of pairs of records for EM from Amazon-Google dataset where in both subfigures the above record is from Amazon and the below record is from Google. In Figure \ref{EM_examples}(a), both records refer to the same real-world entity \textit{adobe photoshop 4.0} although in one record the manufacturer value is missing, and the prices are different. In Figure~\ref{EM_examples}(b), the difference in the value of the \textit{title} attribute in both records clearly indicates that records refer to different entities.

In the past few years, deep learning (DL) has led to a significant improvement in multiple tasks, where DL-based methods achieved state-of-the-art (SOTA) results for text, image, and speech data. In many cases, DL models are trained end-to-end to automatically extract features and build predictive models. This significantly reduces the human effort that is needed in traditional methods for feature engineering, and gives the model the ability to capture specific features that are better than the hand-crafted ones for multiple tasks. Following the success of DL models, researchers have focused on exploring DL in data cleaning and integration. In particular, multiple DL methods have been proposed to solve the EM task \cite{ebraheem_vldb,mpm,active_learning_em,deepmatcher,auto-em}. Based on the DL architecture, we can distinguish two groups of models: attribute-level models where a schema matching step is necessary to compare values of corresponding attributes between two records, and record-level models where the records are compared in their entirety. In addition, traditional pretrained embedding, such as Glove \cite{glove} 
and FastText \cite{fasttext}, is used as word embedding in these DL models.
 Deep contextualized language models (DCLM), like BERT \cite{Devlin2019BERTPO}, RoBERTa \cite{Liu2019RoBERTaAR}, and DistilBERT \cite{distilbert} have been recently proposed to solve multiple tasks \cite{wang-etal-2018-glue,sakata2019,Chen2020TableSU,selab_arxiv,survey_doc_retrieval,selab_ijcnn,strubert}. 
Different from traditional word embeddings, the pre-trained neural language models are contextual where the representation of a token is a function of the entire sentence. This is mainly achieved by the use of a self-attention structure called a Transformer \cite{transformer}. 
Building on DCLM, Ditto \cite{ditto} achieved SOTA results in EM.

Although DL methods have led to a significant improvement in the EM task, these models need a huge amount of labeled data for each domain. DL-based models are trained in a supervised setting for each dataset in EM, where a different model is obtained and is fully fine-tuned on a specific dataset. This means that existing models capture the specific signals for each dataset in EM which leads to overfitting on just one dataset. In addition, the knowledge that is learned from one dataset is not explored to better understand the EM task so that the predictions in other datasets can be made with fewer labeled samples.  

In order to overcome the limitations of prior methods, we propose a new method, called \textit{\textbf{D}omain \textbf{A}daptation for \textbf{M}atching \textbf{E}ntities} (DAME), that transfers the task knowledge from multiple source domains to a target domain. Our method presents a new setting for EM where the objective is to capture task-specific knowledge from pretraining our model using multiple source domains, then testing our model on a target domain. In our study, we are interested in two aspects of our model. First, we study the zero-shot learning (ZSL) case of DAME on the target domain. Second, we study the effect of fine-tuning our proposed model on the target domain using different percentages of training data, and we compare our fine-tuned model to SOTA methods. We formulate EM as a mixture of experts with a global shared model \cite{guo_emnlp,expert1,wright_emnlp} where each expert is trained on an individual source domain, and the global model is trained on all domains. Then, we aggregate the features from the experts using a global model-guided attention mechanism.  We train DAME with unsupervised domain adaptation (DA) loss functions \cite{guo_emnlp,wright_emnlp} to reduce the domain shift between the source and target domains.

In summary, we make the following contributions: (1) We propose a new DA-based method for EM. Our new formulation of EM is based on the mixture of experts where we transfer learning from multiple source domains to a target domain. (2) We study the ZSL case on the target domain and demonstrate that our method learns the EM task and transfers the task knowledge to the target domain. (3) We extensively study fine-tuning our model on the target dataset from multiple domains, and demonstrate that our model generalizes better than SOTA methods for most of the datasets. 

\section{Related work}

\subsection{Entity matching} 
EM \cite{deepmatcher,ditto,active_learning_em,ebraheem_vldb,em_survey} is the field of research that solves the problem of finding records that refer to the same real-world entity. EM, also known as data matching, record linkage, entity resolution, etc, has been intensively studied in recent years because EM is an important step in data cleaning and integration. Given two collections of records $D_1$ and $D_2$, EM classifies a pair of entities $(e_1,e_2), \forall e_1 \in D_1, e_2 \in D_2$ into match or non-match. The records from $D_1$ and $D_2$ can have the same or different set of attributes. The value of each attribute is composed of a sequence of tokens.

Comparing all record pairs from $D_1$ and $D_2$ grows quadratically and it becomes very time-consuming to predict the matching records for the input datasets. 
 Therefore, a set of candidate pairs $C \subset D_1 \times D_2$, where $|C| \ll|D_1 \times D_2|$ is selected in a separate step, called blocking, before running a computationally expensive algorithm for EM. Multiple blocking methods have been proposed in the literature \cite{blocking1,blocking2,blocking3}.
After the blocking step, each record pair $(e_1,e_2) \in C$ is compared to predict a binary label indicating a match or non-match. Prior works have proposed string similarity-based methods to compare records \cite{string_similarity1,string_similarity2,string_similarity3}. Traditional supervised classifiers , such as decision trees, support vector machines, and naive Bayes,
 have been proposed to map the string similarities-based feature vector to a binary class label \cite{classifier1,classifier2}. In addition, rule-based methods have been proposed to solve EM \cite{rule1,rule2,rule3}. 
Recently, DL-based methods have been proposed to solve EM \cite{ebraheem_vldb,mpm,active_learning_em,deepmatcher,auto-em,ditto}. The DL methods of EM can be categorized into attribute- and record-level comparison methods. Attribute comparators predict the label of a pair of records based on the signals collected from matching values of the same attribute. DeepMatcher \cite{deepmatcher}, which is the SOTA attribute-level comparator, explores multiple techniques to compute the attribute representation from word embedding, where combining both bidirectional GRU and decomposable attention \cite{decom_attention} leads to the best results. FastText \cite{fasttext} is used for word embedding in DeepMatcher.
The SOTA method in EM is a record-based comparator known as Ditto \cite{ditto} which is based on DCLM. Ditto models each record by alternating between attributes and data values with two additional special tokens [COL] and [VAL]. Incorporating attribute names in the record representation provides the Transformer \cite{transformer} layers with more information to match attributes of two records. 
 Then, Ditto adapts the sentence pair classification setting to EM in order to compare record pairs using the special tokens [SEP] and [CLS] that are added into the input. In addition, Ditto explores domain-specific optimizations by injecting domain knowledge into the input in the form of span typing and normalization. Ditto uses data augmentation techniques during the training phase with span-, attribute-, and record-level operators consisting of deletion, shuffling, and swapping.

\subsection{Domain adaptation}

DA studies the transfer of task knowledge from a single or multiple labeled source domains to an unlabeled target domain. In this paper, we are interested in the case of multiple source domains known as Multi-Source DA (MSDA). Using only unlabeled data from the target domain is known as Unsupervised DA (UDA).

Existing approaches in UDA focus on reducing the domain shift between the source and target domains by aligning feature vectors \cite{da1,da2}. Representation learning methods have been proposed for UDA such as  domain adversarial networks \cite{adv1,adv2}. and denoising autoencoders \cite{autoencoder}. 
 Other representation learning methods include comparing the marginal distribution between the source and target domains in an adversarial way \cite{guo_emnlp} and minimizing the covariance between the source and target representations \cite{covariance}. An effective strategy in the case of MSDA is known as a mixture of experts \cite{guo_emnlp,expert1,wright_emnlp}. Kim et al. \cite{expert1} proposed to incorporate an attention mechanism to combine the predictions from multiple models trained on the source domains. Guo et al. \cite{guo_emnlp} proposed a method that is based on a mixture of experts where the posteriors of the models are combined using a point-to-set Mahalanobis distance metric between an input sample and source domains.  
Wright and Augenstein \cite{wright_emnlp} improved the performance of the mixture of experts using DCLM as experts in source domains. This work follows a line of research that investigates the use of Transformers \cite{transformer} in DA \cite{gururangan_acl,han_emnlp,ma_emnlp,rietzler_lrec}. Ma et al. \cite{ma_emnlp} improved the performance of BERT in the target domain for natural language inference by incorporating a similarity of a given target domain to source domains with curriculum learning \cite{curriculum_learning}. AdaptaBERT \cite{han_emnlp} is a BERT-based model that is proposed in the case of UDA for the sequence labeling by adding a masked language modeling in the target domain. Fine-tuning of BERT on the target domain was also shown to be effective in the  sentiment analysis task \cite{rietzler_lrec}.  Gururangan et al. \cite{gururangan_acl} combines both domain and task adaptive pretraining to improve the performance of RoBERTa on multiple 
 NLP tasks. The task-adaptive pretraining represents pretraining on unlabeled datasets that are relevant to the task by continuing pretraining RoBERTa on these datasets.

\section{Problem statement}
Our formulation of DA in EM task is based on the unsupervised multi-source DA setting which consists of $K$ labeled source domains $\left\{\mathcal{S}_{i}\right\}_{i=1}^{K}$, where $
\mathcal{S}_{i} =\left\{\left(x_{j}^{\mathcal{S}_{i}}, y_{j}^{\mathcal{S}_{i}}\right)\right\}_{j=1}^{\left|\mathcal{S}_{i}\right|}$ ($x_{j}^{\mathcal{S}_{i}}$ is the $j$-th instance of $\mathcal{S}_{i}$ with a label $y_{j}^{\mathcal{S}_{i}}$), and unlabeled target domain $\mathcal{T}=\left\{x_{j}^{\mathcal{T}}\right\}_{j=1}^{|\mathcal{T}|}$ ($x_{j}^{\mathcal{T}}$ is the $j$-th instance of $\mathcal{T}$). The objective is to learn a classifier $M$ using labeled data from source domains and unlabeled data from the target domain so that (1) $M$ produces accurate predictions on the target domain without fine-tuning (ZSL case), and (2) $M$ generalizes better than SOTA methods on the target domain after partially or fully fine-tuning. 

\section{Domain adaptation for matching entities}

In this section, we introduce our proposed method DAME which is a DA-based method for matching entities. We first describe the architecture of DAME, and then present the DA-based training strategy to update the parameters of our proposed model. Finally, we present our fine-tuning strategy in the case of using labeled samples from the target domain to update DAME. 

\subsection{DAME architecture}

There are multiple datasets that are available for the EM task. Therefore, our model is based on formulating the EM as a mixture of domain experts in the case of DA. Each expert model is trained on one source domain. We denote by $f_{S_i}$, the expert model that is trained on $S_i$. Training a mixture of experts and shared models improves the performance when multiple source domains are available as shown in prior works \cite{guo_emnlp,expert1,wright_emnlp}. Therefore, we also add a global model $g$ that is trained using all the source domains $\left\{\mathcal{S}_{i}\right\}_{i=1}^{K}$. 

DCLM have been proposed in the DA setting to solve multiple tasks \cite{wright_emnlp,gururangan_acl,han_emnlp,ma_emnlp,rietzler_lrec}. We propose to incorporate DCLM in our DA-based model to solve the EM task. Each $f_{S_i}$ and $g$ are initialized using DistillBERT \cite{distilbert} which is a distilled version of BERT with fewer parameters.
We choose to use DistilBERT as the main component for the expert and global models for two reasons. First, by incorporating DCLM, we compare records in their entirety which has been shown to be more effective than attribute-based comparisons. Second, DistilBERT has a reduced size and comparable performance to BERT, and our objective is to include many source domains while keeping the time and memory complexity reasonable. 
In general, our proposed model $M$ has four modules: 
\begin{equation}
M=N\circ Att\circ F \circ Rep
\end{equation}
 $Rep$ is a representation module that produces the sequence input from a pair of records $x$, $F$ is a feature extractor that produces multiple embeddings for the sequence input of the record pair $x$ using expert models $\left\{f_{S_i}\right\}_{i=1}^{K}$ and the global model $g$, $Att$ is an attention module that aggregates the embeddings of the expert models to produce the final multi-source embedding, and $N$ is a classification layer that maps the final embedding to a confidence score to make a matching/non-matching decision on a record pair.

\subsubsection{Representation module $Rep$}

Each record pair $x=(e_1,e_2)$ is composed of two data entries $e_1 \in D_1$ and $e_2 \in D_2$ that correspond to candidate rows from two collections of data entries $D_1$ and $D_2$. \textbf{Both $D_1$ and $D_2$ are from the same source domain}. Each data entry $e_i=\left\{\left(\mathrm{attr}_{j}, \mathrm{val}_{j}\right)\right\}_{1 \leq j \leq C}$ is a set of attribute-value pairs denoted by $(\mathrm{att}_j,\mathrm{val}_j)$, where $C$ is the number of attributes in each record.  We follow the encoding of Ditto \cite{ditto} for serializing data entries to produce a sequence for each record from the attribute-value pairs:
\begin{equation}
r_{e_i}=[\mathrm{COL}] \mathrm{attr}_{1}[\mathrm{VAL}] \mathrm{val}_{1} \ldots[\mathrm{COL}] \mathrm{attr}_{C}[\mathrm{VAL}] \mathrm{val}_{C}
\end{equation}
where $[\mathrm{COL}]$ and $[\mathrm{VAL}]$ are special tokens that denote the start of attributes and values, respectively. The input of EM is a pair of records $x=(e_1,e_2)$. So, $Rep$ takes as input a pair of records, and produces a sequence pair of serialized 
 entries that is given by:
\begin{equation}
Rep(x)=Rep((e_1,e_2))=[\mathrm{CLS}] r_{e_1}[\mathrm{SEP}]r_{e_2}[\mathrm{SEP}],
\end{equation}
where [SEP] and [CLS] are BERT special tokens that are added into the 
 sequence similar to the sentence pair classification setting.

\subsubsection{Feature extractor $F$}

We have $K+1$ DistilBERT models: $K$ expert models $\left\{f_{S_i}\right\}_{i=1}^{K}$ and a global shared model $g$. We use $Rep(x)$ as input to the $K+1$ models to extract $K$ source domain-based embeddings denoted by $f_{S_i}(Rep(x)), i=1,\ldots,K$, and a global model-based embedding denoted by $g(Rep(x))$. The embeddings from the source domain models and the 
 global model are extracted using the hidden state of the [CLS] token from the last Transformer block in each DistilBERT model. In conclusion, the output of $F$ is given by:
\begin{equation}
    F(Rep(x))=\left\{f_{S_i}(Rep(x))\right\}_{i=1}^{K}\cup g(Rep(x))
\end{equation}

\subsubsection{Attention module $Att$}

When aggregating the embeddings that are extracted using $F$, the embeddings from the source domains and the global model should not be treated equally as there are domains that are more relevant to a given record pair $x$ than others. 
We use a parameterized attention model that attends to all domains using a dot product-based attention where three parametric matrices are introduced: a query matrix $ Q\in \mathbb{R}^{d \times d}$, a key matrix $K_e \in \mathbb{R}^{d \times d}$, and a value matrix $V \in \mathbb{R}^{d \times d}$, where $d$ is the dimension of the embedding. We first concatenate all the expert embeddings from $F(Rep(x))$ to form an embedding matrix denoted by $E \in \mathbb{R}^{K \times d}$. The attention operations are defined by:
\begin{equation}
\begin{array}{l}
\alpha=g(Rep(x))^TQ \in \mathbb{R}^{1 \times d} \\
\mathcal{K}=EK_e \in  \mathbb{R}^{K \times d}  \\
\mathcal{V}=EV \in  \mathbb{R}^{K \times d} \\
Att(Rep(x),Q, K, V)=\operatorname{softmax}\left(\frac{\alpha \mathcal{K}^{T}}{\sqrt{d}}\right) \mathcal{V} \in \mathbb{R}^{1 \times d}
\end{array}
\end{equation}
An important design choice in our attention module $Att$ is the use of the global representation $g(Rep(x))$ to map the query matrix $Q$ to a query vector $\alpha$. Given that the global model is trained on all the source domains, we expect the global model's embedding to transfer to the target domain, and by consequence we obtain more accurate attention weights in the target domain to aggregate the source domains, mainly in the zero-shot learning case. The output of the attention module is used as input to the classification layer $N$ to predict the matching score of the input record pair $x$. 

\subsection{Training strategy}

In the multi-source DA setting, we have $K$ labeled source domains $\left\{\mathcal{S}_{i}\right\}_{i=1}^{K}$, where $
\mathcal{S}_{i} =\left\{\left(x_{j}^{\mathcal{S}_{i}}, y_{j}^{\mathcal{S}_{i}}\right)\right\}_{j=1}^{\left|\mathcal{S}_{i}\right|}$, and an unlabeled target domain $\mathcal{T}=\left\{x_{j}^{\mathcal{T}}\right\}_{j=1}^{|\mathcal{T}|}$. Our training phase is based on the multi-task learning setting. In each batch for the training phase, we sample $B$ pairs of records $X_j=(x_1^{\mathcal{S}_{j}},y_1^{\mathcal{S}_{j}}),(x_2^{\mathcal{S}_{j}},y_2^{\mathcal{S}_{j}}),\ldots,(x_B^{\mathcal{S}_{j}},y_B^{\mathcal{S}_{j}})$ from a given source $S_j$. Our loss function $\mathcal{L}$ is composed of four parts and is given by:
\begin{equation}
    \mathcal{L}(X_j)=\lambda_1 \mathcal{L}_{1}(X_j)+\lambda_2 \mathcal{L}_{2}(X_j)+\lambda_3 \mathcal{L}_{3}(X_j)+\lambda_4 \mathcal{L}_{4}(X_j)
\end{equation}
where $\lambda_1$, $\lambda_2$, $\lambda_3$, and $\lambda_4$ are hyperparameters that control the contribution of each loss to the final loss function $\mathcal{L}$; each of $\mathcal{L}_{1}$, $\mathcal{L}_{2}$, $\mathcal{L}_{3}$, and $\mathcal{L}_{4}$ represents a task-specific loss.

\subsubsection{Expert domain loss $\mathcal{L}_{1}$}
$f_{S_i}$ represents the expert model of $S_i$, for all $i \in 1,2,\ldots,K$. To optimize each expert model $f_{S_i}$, we add a classification layer $N_{S_i}$ that predicts the probabilities of matches and non-matches for each domain $S_i$. So, in total we add $K$ classification layers. Given that $X_j$ is sampled from the $j$-th domain, the domain expert loss $\mathcal{L}_{1}$ is given by:
\begin{equation}
    \mathcal{L}_{1}(X_j)= \dfrac{1}{B} \sum_{l=1}^{B} CrossEnt(N_{S_j}(f_{S_j}(Rep(x_l^{\mathcal{S}_{j}}))),y_l^{\mathcal{S}_{j}})
\end{equation}
where $CrossEnt$ denotes the cross entropy loss function. 

\subsubsection{Global model loss $\mathcal{L}_{2}$}

The global model is trained on all the source domains in order to learn a universal embedding for the EM task that supports transfer to the target domain while maintaining important matching signals for each source domain. In addition, the embedding of the global model is multiplied with the query matrix $Q$ in the attention module $Att$ to compute the contribution of each source domain to the final representation. After learning how to aggregate features in the training phase on source domains, the global model guides the attention module $Att$ to pick the most important source domains for the target domain during the testing phase.
 To optimize the global model $g$, we add a classification layer $N_g$ that predicts the probabilities of matches and non-matches for all source domains. 
 The global model loss $\mathcal{L}_{2}$ is given by:
\begin{equation}
    \mathcal{L}_{2}(X_j)= \dfrac{1}{B} \sum_{l=1}^{B} CrossEnt(N_{g}(f_{g}(Rep(x_l^{\mathcal{S}_{j}}))),y_l^{\mathcal{S}_{j}})
\end{equation}

\subsubsection{Meta-target loss $\mathcal{L}_{3}$}

In DA, the objective is to incorporate multiple source domains to predict labels for samples from the target domain during the testing phase. In order to simulate the process of DA during the training phase, we use the meta-target and meta-sources similar to Guo et al. \cite{guo_emnlp}. Given that $X_j$ is sampled from the $j$-th domain, the meta-target is the $j$-th source domain and the meta-sources are $\left\{\mathcal{S}_{i}\right\}_{i=1, i \neq j}^{K}$. The meta-model $M_{S_j}$ differs only on the feature extractor part $F_{S_j}$ compared to $M$. $M_{S_j}$  is given by:
\begin{equation}
M_{S_j}=N\circ Att\circ F_{S_j} \circ Rep
\end{equation}
where:
\begin{equation}
    F_{S_j}(Rep(x))=\left\{f_{S_i}(Rep(x))\right\}_{i=1, i \neq j}^{K}\cup g(Rep(x))
\end{equation}
The same attention module $Att$ is applicable to the output of the meta-feature extractor $F_{S_j}$ where the query matrix based on the global model attends to all the expert embeddings in the key matrix regardless of the number of expert models. Finally, the meta-target loss $\mathcal{L}_{3}$ for the batch $X_j$ is given by:
\begin{equation}
    \mathcal{L}_{3}(X_j)= \dfrac{1}{B} \sum_{l=1}^{B} CrossEnt(M_{S_j}(x_l^{\mathcal{S}_{j}}),y_l^{\mathcal{S}_{j}})
\end{equation}

\subsubsection{Adversarial loss $\mathcal{L}_{4}$}

The global model $g$ plays an important role in the attention module $Att$. Learning a domain invariant embedding from the global model makes the transfer to the target domain smoother as the attention weights should be more accurate. To obtain a domain invariant representation from $g$, we adapt the domain adversarial training for EM. Similar to the generative adversarial network (GAN), a min-max objective function is introduced to optimize the parameters of the generator which is the global model $g$ and the discriminator denoted by $D$. The parameters of $D$ are optimized to predict the domain of a sample $x$ using $g(Rep(x))$, and the parameters of $g$ are optimized to produce a confusing representation $g(Rep(x))$ for $D$. We alternate between updating $D$ and $g$. Given that $X_j$ is sampled from the $j$-th domain, in order to update $D$, we minimize $\mathcal{L}_{D}$ which is given by:
\begin{equation}
    \mathcal{L}_{D}(X_j)= \dfrac{1}{B} \sum_{l=1}^{B} CrossEnt(D(f_{g}(Rep(x_l^{\mathcal{S}_{j}}))),j)
\end{equation}
$\mathcal{L}_{D}$ is minimized with respect to only the parameters of $D$.
Then, we set $\mathcal{L}_{4}(X_j)=-\mathcal{L}_{D}(X_j)$ to update the parameters of $g$ when minimizing $\mathcal{L}$ ($D$ parameters are fixed). Unlabeled samples $\mathcal{T}=\left\{x_{j}^{\mathcal{T}}\right\}_{j=1}^{|\mathcal{T}|}$ from the target domain can also be considered as an additional domain when updating the parameters of $D$ and $g$ by alternating between minimizing $\mathcal{L}_{D}$ and $-\mathcal{L}_{D}$, respectively. In this case, the total number of labels that are used in $\mathcal{L}_{D}$ is equal to $K+1$.

\subsection{Fine-tuning DAME on the target domain}

 During fine-tuning DAME on the target domain, we only update the weights of the global model $g$, attention weights $Att$, and the classification layer $N$, and we keep the weights of the expert models $f_{S_1},f_{S_2},\ldots,f_{S_K}$ frozen. The objective of the fine-tuning step is to slightly update the parameters of DAME to incorporate dataset-specific signals related to the target domain without changing the parameters of expert models. There are multiple fine-tuning scenarios on the target domain. First, we can use all the samples from the target domain or only a limited budget of samples for fine-tuning. Second, in the case of having access to only a limited budget of samples, we can randomly choose samples, or adapt active learning (AL) selection strategies to select the most promising samples. We experiment with all the scenarios and produce AL results using methods from \cite{least_entropy,usde_bald,coreset}.

\section{Evaluation} \label{eval}

\subsection{Data collections}

Table \ref{datasets} represents all the 12 datasets that we use in our experiments. Datasets are collected from the entity resolution Benchmark datasets \cite{er_datasets} and the Magellan data repository \cite{magellan}. These datasets cover multiple domains including clothing, electronics, citation, restaurant, products, music, and software. Each dataset is composed of candidate pairs of records from two structured tables that have the same set of attributes. The datasets vary in the size and this simulates real-world scenarios where there are some domains that are more frequent than others. The total number of attributes in all datasets ranges from 1 to 8. The rate of matches in all datasets ranges from 9.39\% to 24.48\%. Clearly, there is a class imbalance in all datasets where the non-matching class is significantly larger than the matching class. Each dataset is split into training, validation, and testing, and we use the same pre-splited datasets in Ditto \cite{ditto}.

\begin{table}
\centering
\footnotesize
\caption{Datasets for our experiments.}
\vspace*{-2.55mm}
\begin{tabular}{@{}ccccc@{}}
\toprule
Dataset & Domain&Size&\% matches&nb attributes \\
\midrule
Shoes&clothing& 5,805&21.95&1\\\midrule
Cameras&electronics& 5,255&22.03&1\\\midrule
Computers&electronics& 8,094&22.42&1\\\midrule
Watches&electronics& 6,413&22.85&1\\\midrule
DBLP-GoogleScholar&citation&28,707&18.62&4\\\midrule
DBLP-ACM&citation&12,363&17.95&4\\\midrule
Fodors-Zagats&restaurant&946&11.62&6\\\midrule
Beer&product&450&15.11&4\\\midrule
iTunes-Amazon&music&539&24.48&8\\\midrule
Abt-Buy&product&9,575&10.73&3\\\midrule
Amazon-Google&software&11,460&10.18&3\\\midrule
Walmart-Amazon&electronics& 10,242&9.39&5\\

 \bottomrule
\end{tabular}\label{datasets}
\end{table}

\begin{figure*}
        \begin{subfigure}[b]{0.245\textwidth}
                \includegraphics[width=\linewidth]{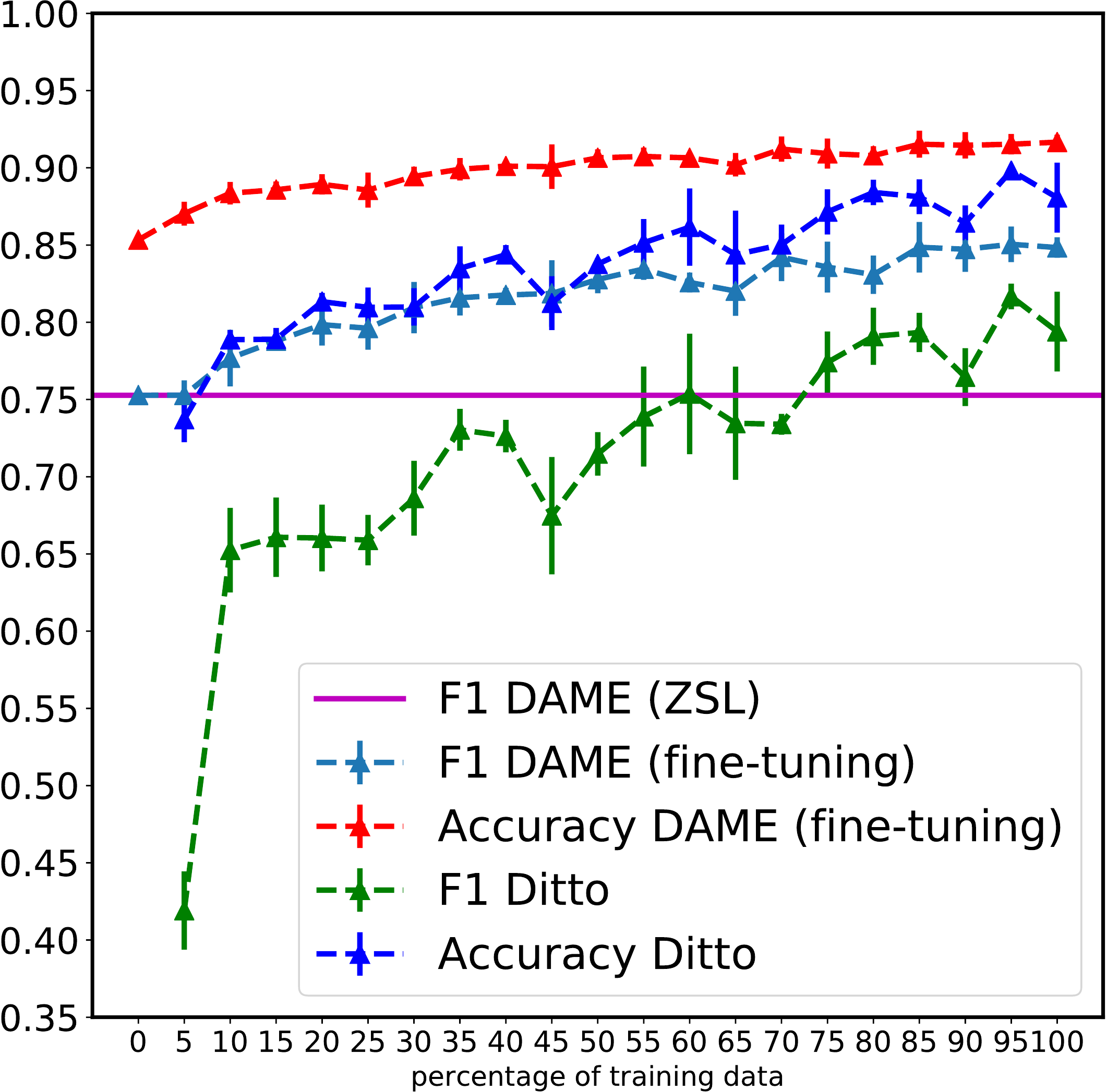}
                \caption{Shoes}
                
        \end{subfigure}%
        \hspace{\fill}
        \begin{subfigure}[b]{0.245\textwidth}
                \includegraphics[width=\linewidth]{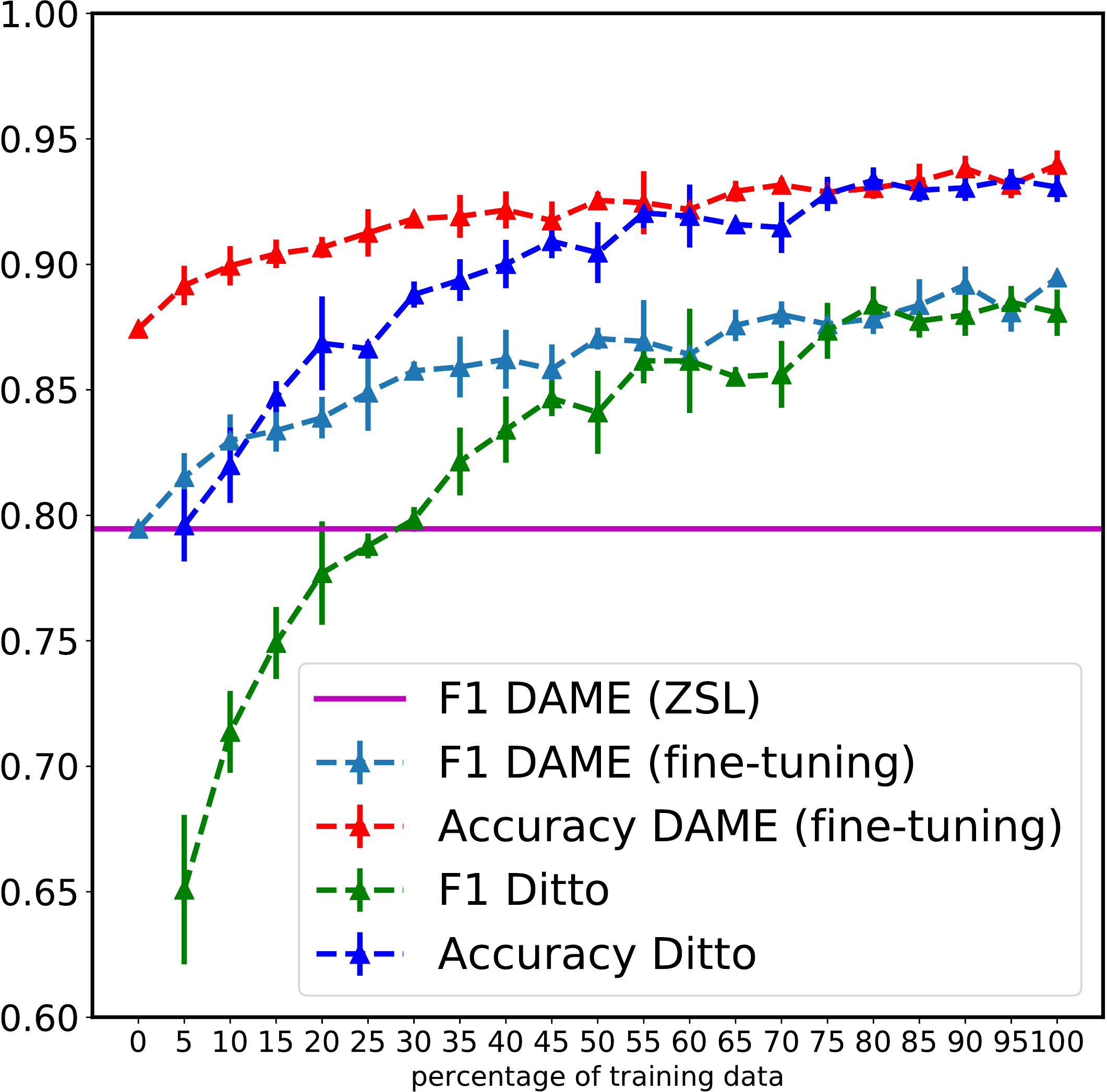}
                \caption{Computers}
               
        \end{subfigure}%
        \hspace{\fill}
        \begin{subfigure}[b]{0.245\textwidth}
                \includegraphics[width=\linewidth]{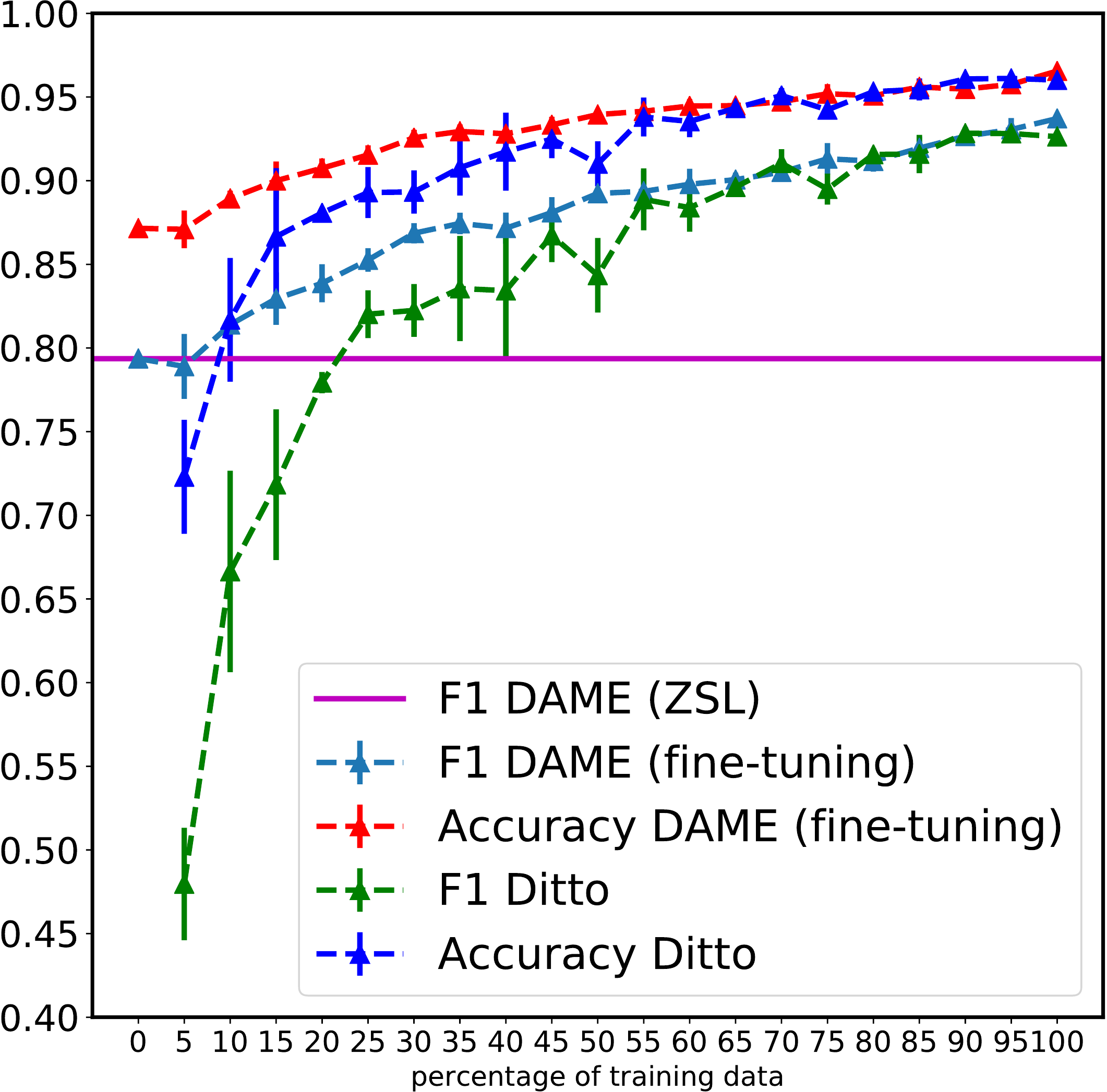}
                \caption{Watches}
               
        \end{subfigure}%
        \hspace{\fill}
        \begin{subfigure}[b]{0.245\textwidth}
                \includegraphics[width=\linewidth]{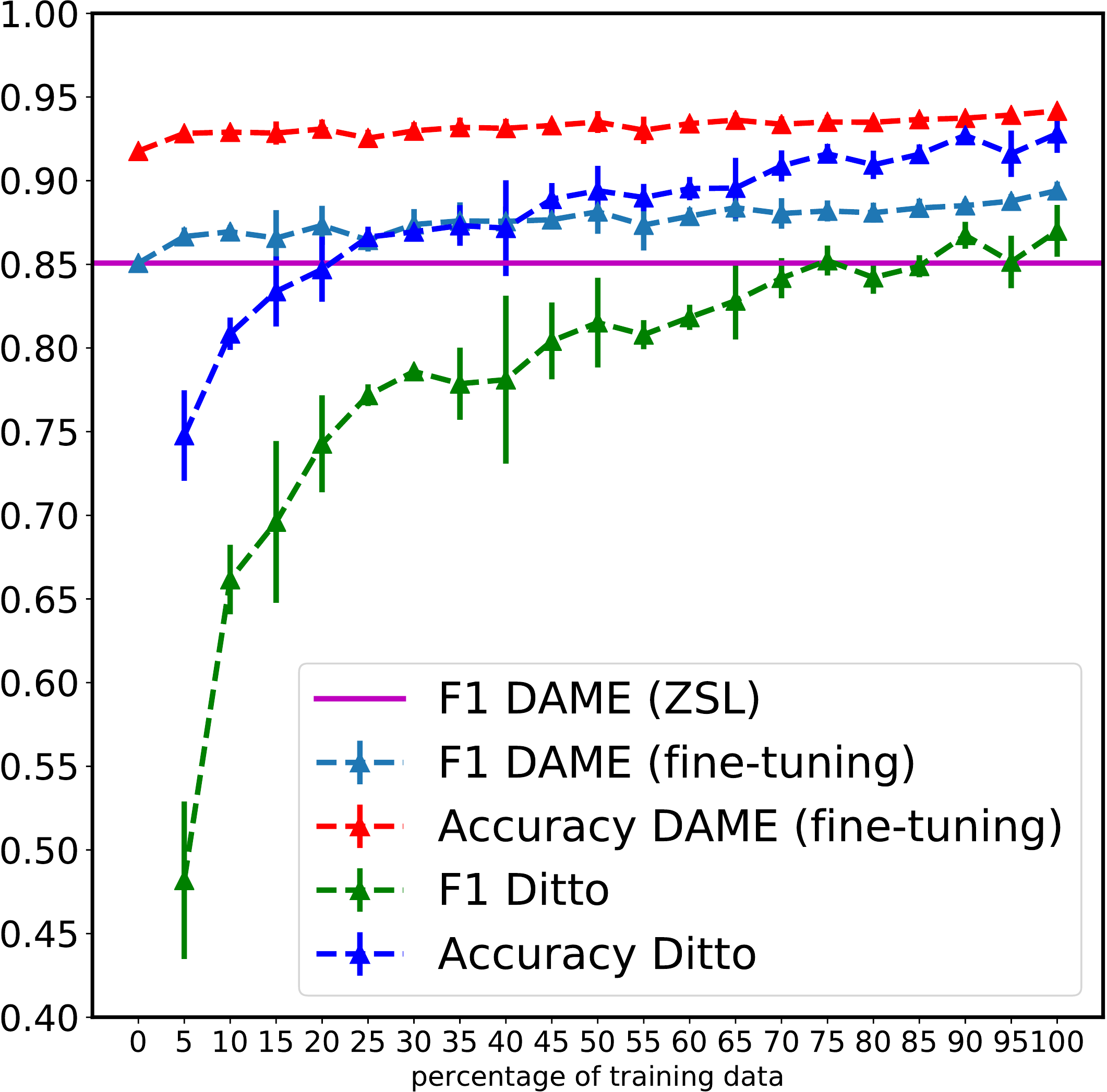}
                \caption{Cameras}
                
        \end{subfigure}
        \vspace*{-3.25mm}
        \caption{Comparison of DAME results against Ditto for datasets with similar structures (Shoes, Computers, Watches, and Cameras). The plots report two evaluation metrics: F1 score and accuracy. In all figures, the light blue plot represents the F1 score of DAME, and is compared against the green plot that represents the F1 score of Ditto; the red plot represents the accuracy of DAME, and is compared against the blue plot that represents the accuracy of Ditto; the magenta color represents the F1 score of the ZSL for the target domain, which is equivalent to 0\% of supervised training data from the target domain.}\label{zero_shot_first_set}
\end{figure*}

\begin{table*}[t!]
\centering
\small
\begin{subtable}[t]{0.48\textwidth}
\begin{tabular}{@{}lllll@{}}
\toprule
Method Name & Precision & Recall & F1 &Accuracy  \\ \midrule
 DeepMatcher \cite{deepmatcher}&\textbf{0.9489}  &0.9373  &0.9431 &0.9789 \\\midrule
 Ditto \cite{ditto}&0.9358  &0.9542  & 0.9449&0.9793  \\\midrule
 DAME (ZSL)&0.9098  & 0.8579 &0.8831 &0.9576  \\\midrule
 DAME (full training data)&0.9354  &\textbf{0.9719} &\textbf{0.9533} & \textbf{0.9850}
 \\ \bottomrule
\end{tabular}
\caption{ DBLP-GoogleScholar}
\label{tab:table1_d}
\end{subtable}
\hspace{\fill}
\begin{subtable}[t]{0.48\textwidth}
\begin{tabular}{@{}lllll@{}}
\toprule
Method Name & Precision & Recall & F1 &Accuracy  \\ \midrule
 DeepMatcher \cite{deepmatcher}&0.9855&0.9869   &0.9861  &0.9945  \\\midrule
 Ditto \cite{ditto}&\textbf{0.9865}  &0.9865  &0.9865 &0.9951  \\\midrule
 DAME (ZSL)&0.8769  & \textbf{0.9954}  &0.9324 &0.9741  \\\midrule
 DAME (full training data)&\textbf{0.9865}  &\textbf{0.9954} &\textbf{0.9909} & \textbf{0.9971}
 \\ \bottomrule
\end{tabular}
\caption{DBLP-ACM}
\label{tab:table1_a}
\end{subtable}
\vspace*{-1mm}
\caption{DA results for EM using datasets with similar structures. (a) the target dataset is DBLP-GoogleScholar and the source dataset is DBLP-ACM; (b) the target dataset is DBLP-ACM and the source dataset is DBLP-GoogleScholar.
}
\label{metrics}
\end{table*}

\subsection{Baselines}

We compare the performance of our proposed model against the best performing method in the category of attribute-level comparators which is DeepMatcher \cite{deepmatcher} (the previous SOTA), and the SOTA in EM which is Ditto \cite{ditto}. We are interested in two aspects of our proposed model DAME. First, we evaluate the ZSL case for DAME by comparing the performance to baselines that are trained on different percentages of training data. Second, we compare the results of fine-tuning DAME on the target domain against training the baselines on the target domain.

\subsection{Experimental Setup}
We evaluate the performance of DAME and baselines on the EM task using precision, recall, F1-score, and accuracy of predictions on the testing set.
We use $\dag$, and $\ddag$ to denote that the difference in a given evaluation metric between Ditto trained on $50\%$ of data and DAME (ZSL) is less than 0.15, and less than 0.1, respectively. We use $\S$ to denote that either the difference between Ditto trained on $50\%$ of data and DAME (ZSL) is less than 0.05 or DAME (ZSL) is better than Ditto trained on $50\%$ of data. 
DAME is trained for 3 epochs on the source domains. We compare fine-tuning results for DAME and baselines after training for 10 epochs on the same percentage of training data from the target domain. The hyperparameters $\lambda_1$, $\lambda_2$, $\lambda_3$, and $\lambda_4$ are fine-tuned for one dataset and then kept the same for all the experiments. We distinguish 3 sets of experiments based on the structure of datasets. The first set of experiments studies DA for Shoes, Cameras, Computers, and Watches. These datasets have a unique attribute which is  \textit{title}. The second set of experiments also studies DA for datasets that have similar structures which are DBLP-GoogleScholar and DBLP-ACM. The set of attributes for these two datasets are \textit{title}, \textit{authors}, \textit{venue}, and \textit{year}. The third set of experiments is related to \textit{DA in the wild} where we study DA using all 12 datasets regardless of the structures and domains. DAME code is available on Github\footnote{https://github.com/medtray/DAME}.

\subsection{Experimental results}
\subsubsection{DA for Shoes, Computers, Watches, Cameras} 

Figure \ref{zero_shot_first_set} shows the comparison of DAME results against Ditto for Shoes, Computers, Watches, and Cameras. The caption of each subfigure represents the target domain, and the remaining 3 domains represent the source domains. Each data point represents the mean of 5 trials, and the vertical line in each data point represents the standard deviation (std). The plots report two evaluation metrics: F1 score and accuracy. In all figures, the light blue plot represents the F1 score of DAME, and is compared against the green plot that represents the F1 score of Ditto; the red plot represents the accuracy of DAME, and is compared against the blue plot that represents the accuracy of Ditto. DAME and Ditto outperform DeepMatcher for all evaluation metrics by a large margin, so that we only include DAME and Ditto results to avoid clutter in the figures. The magenta color represents the F1 score of the DAME (ZSL) for the target domain, which is equivalent to 0\% of supervised training data from the target domain. We achieve high F1 scores for DAME (ZSL) for both Shoes and Cameras datasets, where the F1 score for DAME (ZSL) is equivalent to training Ditto on 72\% and 85\% of training data for the Shoes and Cameras, respectively. The results are lower for Computers and Watches where the F1 score of DAME (ZSL) is equivalent to Ditto trained on around 25\% of training data. Figure \ref{zero_shot_first_set} shows the results of fine-tuning DAME using different percentages of training data. Fine-tuning DAME leads to a better and more stable (smaller std in most fractions of the training data) performance than Ditto for all datasets which means that DAME generalizes better than existing methods in EM for datasets with similar structures. This can be explained by the important role of DA in learning the task so that the weights are better warmed up for EM.

\subsubsection{DA for DBLP-GoogleScholar, DBLP-ACM}

Table \ref{metrics} summarizes the performance of different approaches on the second set of datasets with the same structure which is composed of DBLP-GoogleScholar and DBLP-ACM. In this case, we have one target dataset and one source dataset. We achieve high results for DAME (ZSL) for both datasets. 
 In addition, fine-tuning DAME slightly increases the F1 and accuracy for both datasets. So, consistent with the first set of experiments, we conclude that DAME transfers the task knowledge from the source domains to a target domain in the case of datasets with similar structures.

\subsubsection{DA in the wild} 

We study the case of transferring knowledge between datasets with different domains and structures. We call this setting DA in the wild which simulates real-world scenarios. Table \ref{wild_results} (end of the paper) shows extensive experiments on 12 datasets reporting evaluation metrics for multiple methods. DAME (ZSL) achieves a better F1 score than DeepMatcher trained with 50\% of training data from the target domain for 7 out of 12 datasets.  The difference between the F1 score of Ditto trained on $50\%$ of data and DAME (ZSL) is less than 0.1, and 0.05 for 83\% and 41\%  of datasets, respectively. By comparing the F1 score of fine-tuning all methods using 50\% of training data from the target domain, we achieve SOTA results for 10 out of 12 datasets. By comparing the F1 score of fine-tuning all methods using all training data from the target domain, we achieve SOTA results for 10 out of 12 datasets. This means that DAME generalizes better than existing methods for datasets in the wild.

\begin{figure*}
        \begin{subfigure}[b]{0.3\textwidth}
                \includegraphics[width=1.2\linewidth]{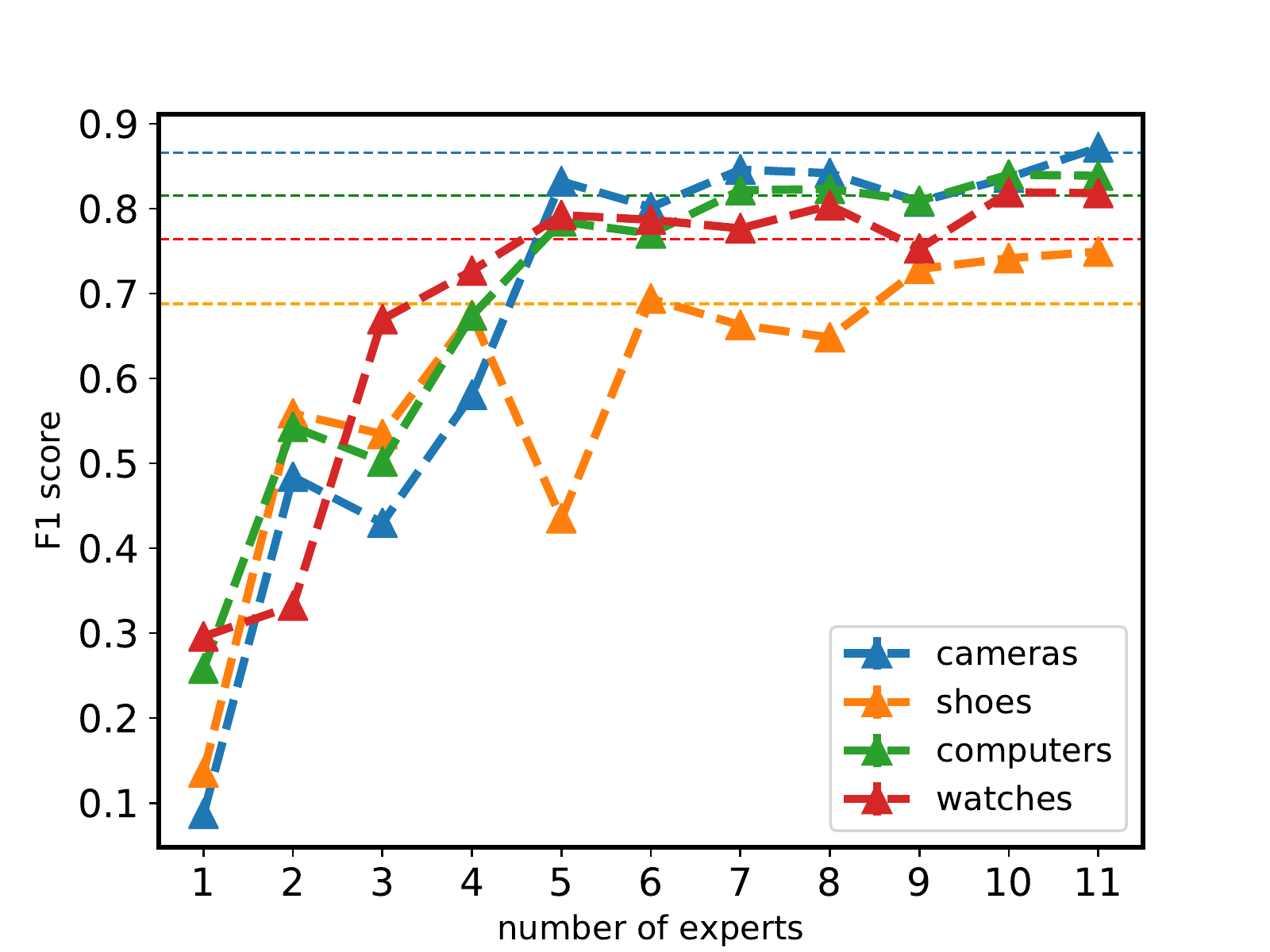}
                \caption{}
                
        \end{subfigure}%
        \hspace{\fill}
        \begin{subfigure}[b]{0.3\textwidth}
                \includegraphics[width=1.2\linewidth]{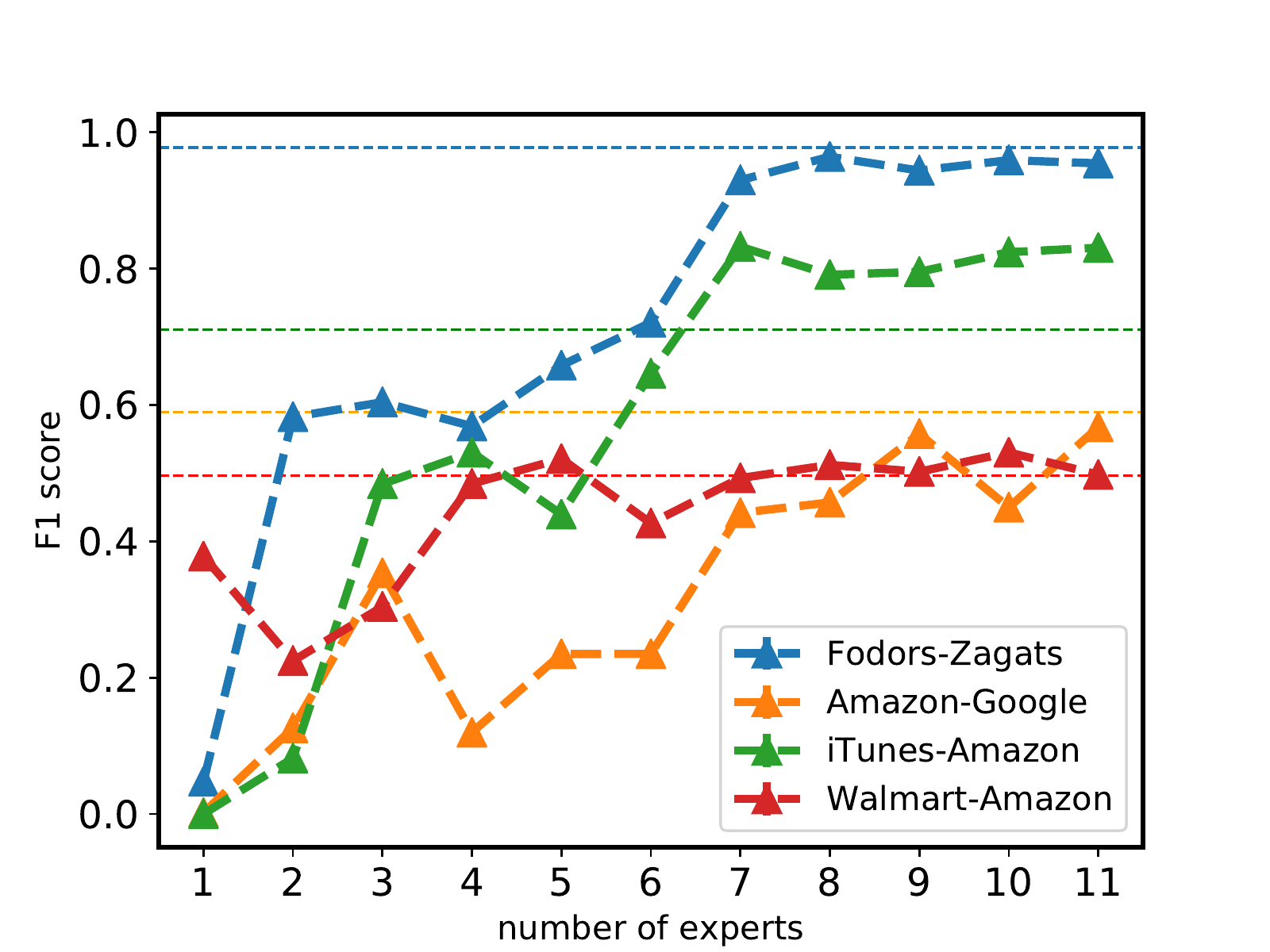}
                \caption{}
               
        \end{subfigure}%
        \hspace{\fill}
        \begin{subfigure}[b]{0.3\textwidth}
                \includegraphics[width=1.2\linewidth]{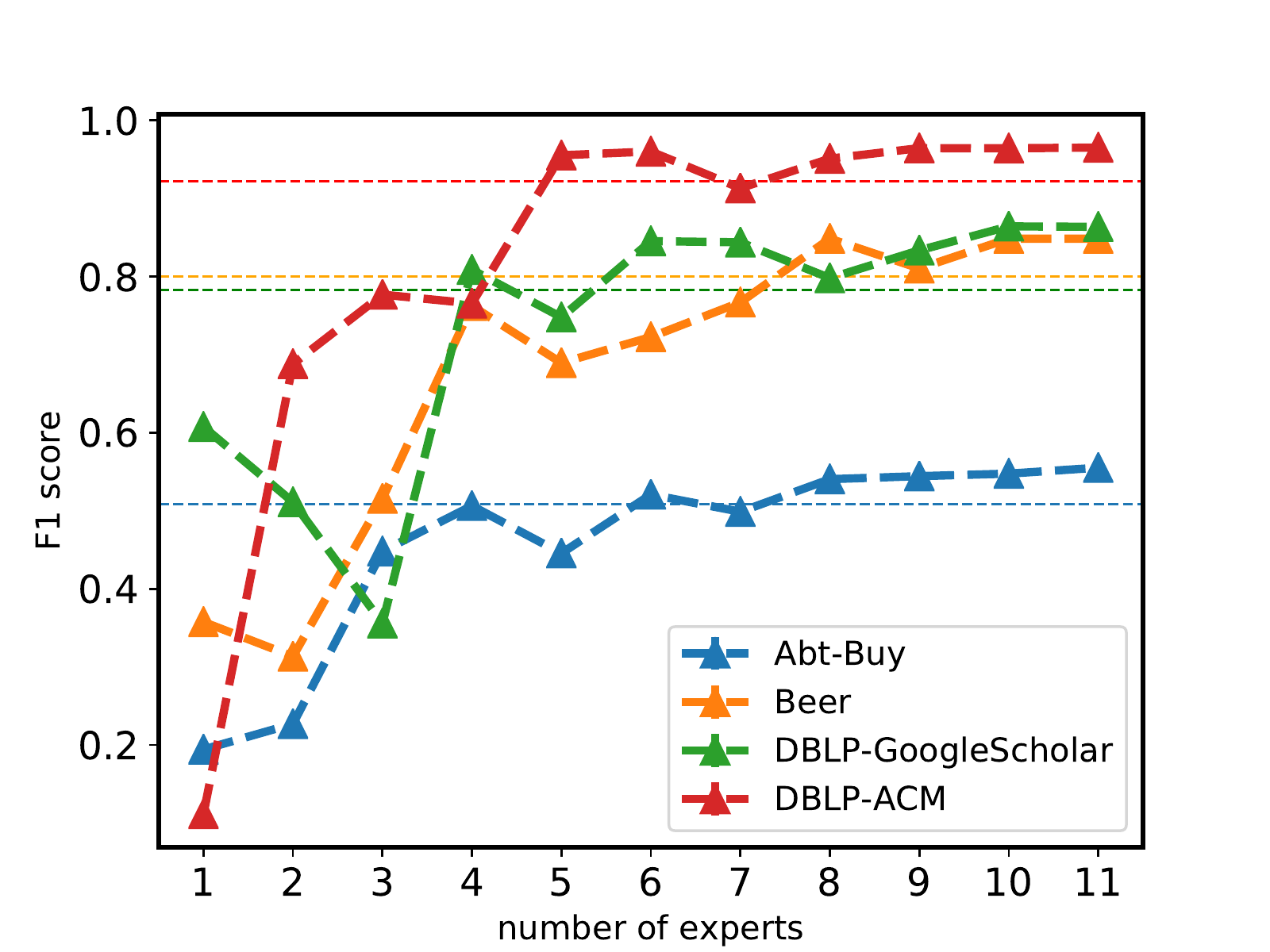}
                \caption{}
               
        \end{subfigure}
        \vspace*{-3.25mm}
        \caption{Comparison of F1 score results with different numbers of expert domains against using global model representation during testing phase on the target domain.}\label{different experts}
\end{figure*}

\begin{figure*}
        \begin{subfigure}[b]{0.245\textwidth}
                \includegraphics[width=\linewidth]{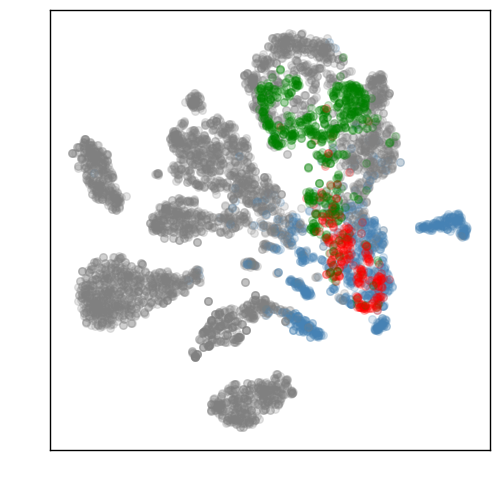}
                \caption{Computers}
                
        \end{subfigure}%
        \hspace{\fill}
        \begin{subfigure}[b]{0.245\textwidth}
                \includegraphics[width=\linewidth]{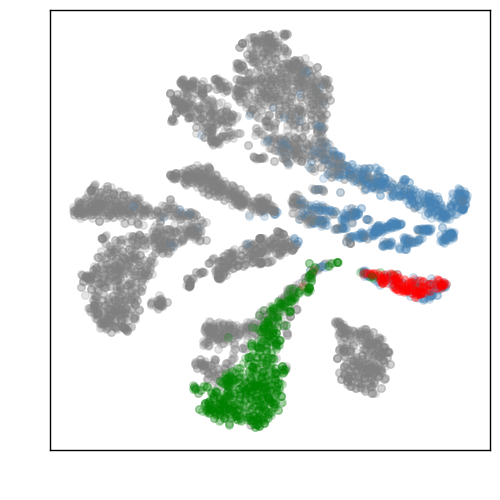}
                \caption{DBLP-ACM}
               
        \end{subfigure}%
        \hspace{\fill}
        \begin{subfigure}[b]{0.245\textwidth}
                \includegraphics[width=\linewidth]{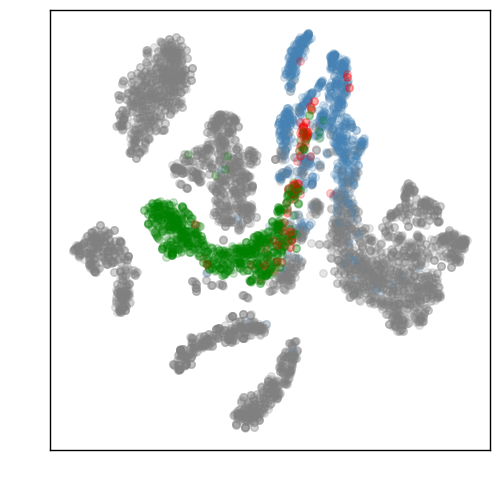}
                \caption{Amazon-Google}
               
        \end{subfigure}%
        \hspace{\fill}
        \begin{subfigure}[b]{0.245\textwidth}
                \includegraphics[width=\linewidth]{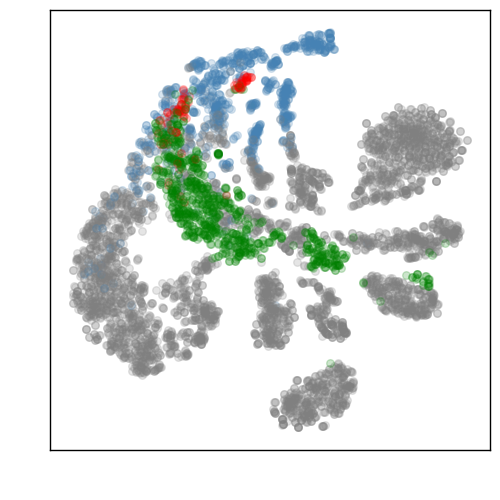}
                \caption{Walmart-Amazon}
                
        \end{subfigure}
        \vspace*{-3.25mm}
        \caption{The t-SNE visualization of the final embeddings for the target and source domains after DA in the wild (ZSL case). The gray and blue colors represent randomly selected data points from the source domains with label 0 and label 1, respectively; The green and red colors represent randomly selected data points from the testing set of the target domain with label 0 and label 1, respectively. 12 domains are used in each experiment, where the caption of each subfigure represents the target domain, and the 11 remaining datasets represent the source domains.}\label{emb_wild}
\end{figure*}

\subsubsection{Expert models vs Global model} 

Figure \ref{different experts} shows the comparison of F1 score results with different numbers of expert domains against using the global model representation during the testing phase on the target domain in the case of ZSL. The x-axis represents the number of experts that we use for predictions. For example, if the number of experts is equal to 6, it means that we randomly choose 6 experts and we drop the remaining 5 experts. Each data point in Figure \ref{different experts} represents an average of 5 trials. The dashed line represents the F1 score for the global model. For 10 out of 12 datasets, combining multiple experts using the attention network $Att$ leads to better results than the global model. Figure \ref{different experts} shows that the fewest number of experts needed to outperform the global model was 5 (DBLP-ACM); the most required was 11 (Cameras). Overall, we obtain better F1 scores for the mixture of experts when we increase the number of experts. This means that the experts help to better understand the EM task, and therefore transfer the learned task knowledge to the 
unseen target domain.

\begin{table}
\centering
\footnotesize
\caption{F1 results for AL after DA.}
\vspace*{-2.25mm}
\begin{tabular}{@{}ccccc@{}}
\toprule
Method & Shoes&Computers&Watches&Cameras \\
\midrule
DAME (ZSL)&0.7527  &0.7946 &0.7936 &0.8507 \\
DAME (full training data)&0.8483&0.8947&0.9371&0.8941\\\midrule
Random Sampling (5\%)&0.7527  &0.8181 &0.8004 &0.8664 \\
Least Confidence \cite{least_entropy} (5\%)&0.7818&0.8402&0.8209&0.8745\\
Entropy Sampling \cite{least_entropy} (5\%)&0.7859&0.8464&0.8166&0.8748\\
USDE \cite{usde_bald} (5\%)&\textbf{0.7877  }&0.8437&0.8151&\textbf{0.8775  }\\
BALD \cite{usde_bald} (5\%)&0.7852&\textbf{0.8472  }&\textbf{0.8313  }&0.8705\\
K-Centers Greedy \cite{coreset} (5\%)&0.7674&0.8271&0.8206&0.8687\\
K-Means \cite{coreset} (5\%)&0.7527&0.8042&0.8097&0.8596\\
Core-Set \cite{coreset} (5\%)&0.7621&0.8304&0.8168&0.8734\\
 \bottomrule
Random Sampling (25\%)&0.8120 &0.8418 &0.8528 &0.8741 \\
Least Confidence \cite{least_entropy} (25\%)&0.8228&0.8804&0.8677&0.8888\\
Entropy Sampling \cite{least_entropy} (25\%)&0.8207&0.8770&0.8740&0.8925\\
USDE (25\%) \cite{usde_bald}&\textbf{0.8286  }&0.8741&0.8688&0.8842\\
BALD (25\%) \cite{usde_bald}&0.8247&\textbf{0.8835  }&\textbf{0.8872  }&\textbf{0.8941 }\\
K-Centers Greedy \cite{coreset} (25\%)&0.8155&0.8771&0.8869&0.8780\\
K-Means \cite{coreset} (25\%)&0.8057&0.8658&0.8694&0.8737\\
Core-Set \cite{coreset} (25\%)&0.8161&0.8696&0.8812&0.8776\\
 \bottomrule
\end{tabular}\label{al_results}
\end{table}

\subsubsection{DAME with Active learning} 

So far, we have discussed the performance of fine-tuning DAME using randomly selected samples from the target domain. To improve the results of fine-tuning our model, we investigate multiple AL selection techniques given a limited budget of labeled instances. Table \ref{al_results} shows the results of multiple AL selection methods applied to the DAME (ZSL) model. The starting point is our DA-based model which is not fine-tuned on the target domain, and the best performance corresponds to DAME fine-tuned on all training data from the target domain. We report results using two budget levels: 5\% and 25\% of the training data from the target domain. The simplest baseline is Random Sampling. The remaining baselines can be categorized into two groups: the confidence-based baselines which are: Least Confidence \cite{least_entropy}, Entropy Sampling \cite{least_entropy}, Uncertainty Sampling with Dropout Estimation (USDE) \cite{usde_bald}, and Bayesian Active Learning Disagreement (BALD) \cite{usde_bald}; and the embedding-based baselines which are  K-Centers Greedy \cite{coreset}, K-Means \cite{coreset}, and Core-Set \cite{coreset}. The selection of samples in the first group is based on the confidence scores of the training data from the target domain that are computed using the DAME (ZSL) model. For example, for a budget of $b$ samples, Least Confidence corresponds to the top $b$ samples with the lowest confidence level. Multiple predictions for a given sample are needed for USDE and BALD to compute the uncertainty functions, and we obtain these different predictions by activating the dropout layers during the inference phase on the target domain. The second group is based on the embeddings of samples from the target domain that are obtained using the DAME (ZSL) model. Clustering of the input space is then applied to determine centers of clusters or core sets. 
Table \ref{al_results} shows that the confidence-based methods lead to better results than the embedding-based methods. In particular, when we select 25\% of samples using the BALD method for the Cameras dataset, we achieve the same F1 score of a fully fine-tuned DAME model using all training data from the target domain. This indicates that the predictions from the classification layer $N$ of our model $M$ accurately reflect the data points where DA was unsuccessful. Therefore, by fine-tuning on these samples from the training data, our model generalizes better on the testing set of the target domain. 

\subsubsection{Visualization}
We show the embedding of DAME in the case of ZSL. Figure \ref{emb_wild} shows the t-SNE visualization of the final embeddings for the target and source domains after DA in the wild (ZSL case). We only show the embeddings of four domains due to the space limitation in the paper, but we notice similar patterns for all the datasets. The gray and blue colors represent randomly selected data points from the source domains with a label 0 and label 1, respectively; the green and red colors represent randomly selected data points from the testing set of the target domain with a label 0 and label 1, respectively. 12 domains are used in each experiment, where the caption of each subfigure in Figure \ref{emb_wild} represents the target domain, and the 11 remaining datasets represent the source domains. The best case is to have a mixture of blue and red dots which represent the matching class for the source and target domains, respectively, and a mixture of gray and green dots which represent the non-matching class for the source and target domains, respectively. This means that we transfer the task knowledge from sources to the target domain for both labels. For example, for Computers and DBLP-ACM, we obtain embeddings that respect the matching and non-matching classes as shown in Figure \ref{emb_wild} (a) and (b), respectively. On the other hand, for Amazon-Google and Walmart-Amazon, there are green dots that are closer to the blue dots than the gray dots as shown in Figure \ref{emb_wild} (c) and (d), respectively, and this leads to incorrect predictions for DAME (ZSL).

\section{Conclusions}
We have shown that our proposed model transfers learning from multiple source domains to an unseen target domain in the EM task. We formulate the EM task as a mixture of experts that capture task-specific knowledge from pretraining on multiple source domains and testing on a target domain. We evaluate DAME in two aspects. First, we study the ZSL case on the target domain and demonstrate that DAME learns the EM task and transfers knowledge to the target domain. Second, we study fine-tuning DAME on the target domain and demonstrate that DAME generalizes better than SOTA methods for most of the datasets. We showed that our results hold in two scenarios which are EM for datasets with similar structures and EM in the wild. Our experimental section contains extensive experiments over 12 datasets with different domains, sizes and structures. In addition, we showed the importance of selecting a specific set of samples in the fine-tuning 
 of the target domain by studying AL methods with limited budget. 
 
Future work includes extending our model to pairs of records with different sets of attributes, and enriching our DA-based model with external knowledge, such as knowledge graphs, to better understand the EM task and therefore transfer more knowledge to the target domain.

\begin{table*}[ht]
\scriptsize
\caption{DA results for EM in the wild. 
\label{trresults1}}
\centering
\begin{tabular}{||c|c|c|c|c|c||}
\hline
\textbf{Target dataset} & \textbf{Method} & \textbf{Precision} & \textbf{Recall} & \textbf{F1} & \textbf{Accuracy} \\ \hline
\multirow{7}{*}{Fodors-Zagats} &DAME (ZSL)  & 0.9565\textsuperscript{\S}  &1.0000\textsuperscript{\S} &0.9777\textsuperscript{\S} &0.9947\textsuperscript{\S}\\\cline{2-6}
& DeepMatcher\cite{deepmatcher} (50\% training data) &0.9360$\pm$0.0559  &0.8333$\pm$0.0428 &0.8801$\pm$0.0334 &0.9735$\pm$0.0074\\&
 Ditto \cite{ditto} (50\% training data)  & \textbf{1.0000}  &0.9545 &0.9767 &\textbf{0.9947}\\&
 DAME (50\% training data)  &0.9565   &\textbf{1.0000} &\textbf{0.9777} &\textbf{0.9947}\\\cline{2-6}&
 DeepMatcher\cite{deepmatcher} (full training data) &0.9092$\pm$0.0756  &0.9848$\pm$0.0214 &0.9437$\pm$0.0423 &0.9858$\pm$0.0108\\&
 Ditto \cite{ditto} (full training data)  &\textbf{1.0000}   &0.9545 &0.9767 &0.9947\\&
 DAME (full training data) &\textbf{1.0000}   &\textbf{1.0000} &\textbf{1.0000} &\textbf{1.0000} \\
 \hline
 \multirow{7}{*}{Beer} &DAME (ZSL)  &0.7368\textsuperscript{\S}   &1.000\textsuperscript{\S} &0.8484\textsuperscript{\S} &0.9450\textsuperscript{\S}\\\cline{2-6}
& DeepMatcher\cite{deepmatcher} (50\% training data) &\textbf{0.8095$\pm$0.0673}  &0.4047$\pm$0.0336 &0.5396$\pm$0.0448 &0.8937$\pm$0.0103\\&
 Ditto \cite{ditto} (50\% training data)  &0.7211$\pm$0.0288   &0.6428 &0.6794$\pm$0.0128 &0.9065$\pm$0.0054\\&
 DAME (50\% training data)  &0.7801 $\pm$0.0433 &\textbf{1.000} &\textbf{0.8758$\pm$0.0273} &\textbf{0.9560$\pm$0.0109}\\\cline{2-6}&
 DeepMatcher\cite{deepmatcher} (full training data)  &0.8183$\pm$0.0581   &0.7142$\pm$0.0583 &0.7588$\pm$0.0193 &0.9304$\pm$0.0051\\&
 Ditto \cite{ditto} (full training data)  &\textbf{0.8174$\pm$0.0396}   &0.9285$\pm$0.0714 &0.8660$\pm$0.0089 &\textbf{0.9560}\\&
 DAME (full training data) &0.7801   &\textbf{1.000} &\textbf{0.8758} &\textbf{0.9560}\\
\hline
\multirow{7}{*}{iTunes-Amazon} &DAME (ZSL)  &0.6750   &1.000\textsuperscript{\S} &0.8059\textsuperscript{\ddag} &0.8807\textsuperscript{\ddag}\\\cline{2-6}
& DeepMatcher\cite{deepmatcher} (50\% training data)  &0.9005$\pm$0.0226   &0.7901$\pm$0.0698 &0.8406$\pm$0.0464 &0.9266$\pm$0.0198\\&
 Ditto \cite{ditto} (50\% training data)  &0.8685$\pm$0.0114   &0.8518$\pm$0.0370 &0.8594$\pm$0.0132 &0.9311$\pm$0.0045\\&
 DAME (50\% training data)  &\textbf{0.9333$\pm$0.0666}   &\textbf{0.9629} &\textbf{0.9467$\pm$0.0344} &\textbf{0.9724$\pm$0.0183}\\\cline{2-6}&
 DeepMatcher\cite{deepmatcher} (full training data)  &0.9139$\pm$0.0149   &0.9135$\pm$0.0174 &0.9135$\pm$0.0088 &0.9571$\pm$0.0043\\&
 Ditto \cite{ditto} (full training data)  &0.9282$\pm$0.0317   &0.9259$\pm$0.0370 &0.9258$\pm$0.0027 &0.9633\\&
 DAME (full training data) &\textbf{0.9807$\pm$0.0192}   &\textbf{0.9259} &\textbf{0.9524$\pm$0.0090} &\textbf{0.9770$\pm$0.0045}\\
\hline
\multirow{7}{*}{Abt-Buy} &DAME (ZSL)  &0.4545   &0.6796\textsuperscript{\dag} &0.5447 &0.8778\textsuperscript{\ddag}\\\cline{2-6}
& DeepMatcher\cite{deepmatcher} (50\% training data)  &0.6978$\pm$0.0416   &0.5355$\pm$0.0397 &0.6033$\pm$0.0112 &0.9244$\pm$0.0021\\&
 Ditto \cite{ditto} (50\% training data)  &0.7916$\pm$0.0312   &0.7839$\pm$0.0169 &0.7870$\pm$0.0069 &0.9543$\pm$0.0028\\&
 DAME (50\% training data)  &\textbf{0.7960$\pm$0.0078}   &\textbf{0.7864$\pm$0.0097} &\textbf{0.7911$\pm$0.0088} &\textbf{0.9553$\pm$0.0018}\\\cline{2-6}&
 DeepMatcher\cite{deepmatcher} (full training data)  &0.7382$\pm$0.0214   &0.6181$\pm$0.0127 &0.6725$\pm$0.0082 &0.9352$\pm$0.0022\\&
 Ditto \cite{ditto} (full training data)  & \textbf{0.9206$\pm$0.0095}  &0.7864$\pm$0.0097 &\textbf{0.8481$\pm$0.0015} &\textbf{0.9697}\\&
 DAME (full training data) &0.8243$\pm$0.0252   &\textbf{0.8592$\pm$0.0097} &0.8410$\pm$0.0084 &0.9650$\pm$0.0026\\
\hline
\multirow{7}{*}{Amazon-Google} &DAME (ZSL)  &0.5431   &0.6453\textsuperscript{\S} &0.5898\textsuperscript{\ddag} &0.9084\textsuperscript{\S} \\\cline{2-6}
& DeepMatcher\cite{deepmatcher} (50\% training data)  &0.5623$\pm$0.0395   &0.5327$\pm$0.0708 &0.5416$\pm$0.0161 &0.9085$\pm$0.0063\\&
 Ditto \cite{ditto} (50\% training data)  &\textbf{0.7055$\pm$0.0037}   &0.6709$\pm$0.017 &\textbf{0.6877$\pm$0.0107} &\textbf{0.9378$\pm$0.0015}\\&
 DAME (50\% training data)  &0.6339$\pm$0.0327   &\textbf{0.7435$\pm$0.0683} &0.6809$\pm$0.0099 &0.9291$\pm$0.0032\\\cline{2-6}&
 DeepMatcher\cite{deepmatcher} (full training data)  &0.7002$\pm$0.0281   &0.6011$\pm$0.0344 &0.6454$\pm$0.0082 &0.9326$\pm$0.0017\\&
 Ditto \cite{ditto} (full training data)  & 0.6709$\pm$0.0077  &\textbf{0.8098$\pm$0.0064} &0.7338$\pm$0.0020 &0.9400$\pm$0.0010\\&
 DAME (full training data) & \textbf{0.7046$\pm$0.0038}  &0.7692$\pm$0.0213 &\textbf{0.7353$\pm$0.0076} &\textbf{0.9435$\pm$0.0006}\\ [1ex]
\hline
\multirow{7}{*}{Shoes} &DAME (ZSL)  &0.6798\textsuperscript{\S}   &0.8135\textsuperscript{\S} &0.7407\textsuperscript{\S} &0.8450\textsuperscript{\S} \\\cline{2-6}
& DeepMatcher\cite{deepmatcher} (50\% training data)  &0.6346$\pm$0.0250   &0.7163$\pm$0.0303 &0.6719$\pm$0.0041 &0.8096$\pm$0.0082\\&
 Ditto \cite{ditto} (50\% training data)  &0.7137$\pm$0.0240   &0.7559$\pm$0.0881 &0.7301$\pm$0.0290 &0.8496$\pm$0.0046\\&
 DAME (50\% training data)  &\textbf{0.8234$\pm$0.0189}   &\textbf{0.8423$\pm$0.0084} &\textbf{0.8325$\pm$0.0055} &\textbf{0.9077$\pm$0.0046}\\\cline{2-6}&
 DeepMatcher\cite{deepmatcher} (full training data)  &0.6908$\pm$0.0366   &0.7988$\pm$0.0162 &0.7400$\pm$0.0179 &0.8468$\pm$0.0158\\&
 Ditto \cite{ditto} (full training data)  &0.7569$\pm$0.0377   &0.8389$\pm$0.0118 &0.7950$\pm$0.0155 &0.8819$\pm$0.0129\\&
 DAME (full training data) &\textbf{0.8421$\pm$0.0222}   &\textbf{0.8796$\pm$0.0152} &\textbf{0.8600$\pm$0.0043} &\textbf{0.9220$\pm$0.0041}\\ [1ex]
\hline
\multirow{7}{*}{Computers} &DAME (ZSL)  &0.7957\textsuperscript{\S}   &0.8729\textsuperscript{\S} &0.8325\textsuperscript{\S} &0.9043\textsuperscript{\S} \\\cline{2-6}
& DeepMatcher\cite{deepmatcher} (50\% training data)  &0.5762$\pm$0.0239   &0.7547$\pm$0.0536 &0.6529$\pm$0.0315 &0.7820$\pm$0.0169\\&
 Ditto \cite{ditto} (50\% training data)  &0.8020$\pm$0.0085   &\textbf{0.9080$\pm$0.0083} &0.8517$\pm$0.0085 &0.9139$\pm$0.0050\\&
 DAME (50\% training data)  &\textbf{0.8303$\pm$0.0268}   &0.9063$\pm$0.0234 &\textbf{0.8659$\pm$0.0039} &\textbf{0.9234$\pm$0.0045}\\\cline{2-6}&
 DeepMatcher\cite{deepmatcher} (full training data)  &0.7002$\pm$0.0258   &0.8350$\pm$0.0356 &0.7614$\pm$0.0270 &0.8576$\pm$0.0157\\&
 Ditto \cite{ditto} (full training data)  &\textbf{0.8682}   &0.9147$\pm$0.0117 &0.8908$\pm$0.0055 &0.9389$\pm$0.0027\\&
 DAME (full training data) &0.8630$\pm$0.0076   &\textbf{0.9264$\pm$0.0033} &\textbf{0.8935$\pm$0.0025} &\textbf{0.9398$\pm$0.0018}\\ [1ex]
\hline
\multirow{7}{*}{Watches} &DAME (ZSL)  &0.7267\textsuperscript{\dag}   &0.9124\textsuperscript{\S} &0.8090\textsuperscript{\ddag} &0.8834\textsuperscript{\ddag}\\\cline{2-6}
& DeepMatcher\cite{deepmatcher} (50\% training data)  &0.6997$\pm$0.0260   &0.7274$\pm$0.0478 &0.7126$\pm$0.0314 &0.8415$\pm$0.0151\\&
 Ditto \cite{ditto} (50\% training data)  &0.8664$\pm$0.0037   &0.8996$\pm$0.0054 &0.8827$\pm$0.0045 &0.9352$\pm$0.0024\\&
 DAME (50\% training data)  &\textbf{0.8691$\pm$0.0196}   &\textbf{0.9160$\pm$0.0109} &\textbf{0.8917$\pm$0.0051} &\textbf{0.9397$\pm$0.0039}\\\cline{2-6}&
 DeepMatcher\cite{deepmatcher} (full training data)  &0.7771$\pm$0.0093   &0.8309$\pm$0.0169 &0.8030$\pm$0.0087 &0.8896$\pm$0.0044\\&
 Ditto \cite{ditto} (full training data)  &\textbf{0.9145$\pm$0.0030}   &0.9178$\pm$0.0164 &0.9161$\pm$0.0097 &0.9545$\pm$0.0049\\&
 DAME (full training data) &0.9010$\pm$0.0038   &\textbf{0.9470$\pm$0.0091} &\textbf{0.9234$\pm$0.0023} &\textbf{0.9575$\pm$0.0009}\\ [1ex]
\hline
\multirow{7}{*}{Cameras} &DAME (ZSL)  &0.8376\textsuperscript{\S}   &0.8958\textsuperscript{\S} &0.8657\textsuperscript{\S} &0.9243\textsuperscript{\S}\\\cline{2-6}
& DeepMatcher\cite{deepmatcher} (50\% training data)  &0.5896$\pm$0.0063   &0.6863$\pm$0.0630 &0.6328$\pm$0.0275 &0.7842$\pm$0.0058\\&
 Ditto \cite{ditto} (50\% training data)  &0.7585$\pm$ 0.0694  &0.8628$\pm$0.0607 &0.8020$\pm$0.0127 &0.8831$\pm$0.0175\\&
 DAME (50\% training data)  &\textbf{0.8801 $\pm$0.0312}  &\textbf{0.8871$\pm$0.0295} &\textbf{0.8825$\pm$0.0011} &\textbf{0.9356$\pm$0.0028}\\\cline{2-6}&
 DeepMatcher\cite{deepmatcher} (full training data)  &0.6986$\pm$0.0280   &0.7847$\pm$0.0075 &0.7388$\pm$0.0159 &0.8486$\pm$0.0124\\&
 Ditto \cite{ditto} (full training data)  &0.8573$\pm$0.0075   &0.9062$\pm$0.0173 &0.8809$\pm$0.0042 &0.9333$\pm$0.0014\\&
 DAME (full training data) &\textbf{0.8917$\pm$0.0013}   &\textbf{0.9070$\pm$0.0017} &\textbf{0.8963$\pm$0.0001} &\textbf{0.9432}\\ [1ex]
\hline
\multirow{7}{*}{Walmart-Amazon} &DAME (ZSL)  &0.3558   &0.9015\textsuperscript{\S} &0.5102 &0.8369\textsuperscript{\dag}\\\cline{2-6}
& DeepMatcher\cite{deepmatcher} (50\% training data)  &0.6938$\pm$0.0171   &0.5474$\pm$0.0217 &0.6118$\pm$0.0167 &0.9346$\pm$0.0023\\&
 Ditto \cite{ditto} (50\% training data)  & \textbf{0.8501$\pm$0.0206}  &0.7098$\pm$0.0466 &0.7721$\pm$0.0191 &0.9607$\pm$0.0017\\&
 DAME (50\% training data)  &0.8082$\pm$0.0019   &\textbf{0.8083$\pm$0.0103} &\textbf{0.8082$\pm$0.0061} &\textbf{0.9638$\pm$0.0009}\\\cline{2-6}&
 DeepMatcher\cite{deepmatcher} (full training data)  &0.6971$\pm$0.0183   &0.6010$\pm$0.0223 &0.6448$\pm$0.0067 &0.9376$\pm$0.0011\\&
 Ditto \cite{ditto} (full training data)  & \textbf{0.8883$\pm$0.0459}  &0.7694$\pm$0.0336 &\textbf{0.8227$\pm$0.0004} &\textbf{0.9687$\pm$0.0014}\\&
 DAME (full training data) &0.8615$\pm$0.0090   &\textbf{0.7875$\pm$0.0207} &0.8226$\pm$0.0071 &0.9680$\pm$0.0007\\ [1ex]
\hline
\multirow{7}{*}{DBLP-GoogleScholar} &DAME (ZSL)  &   0.9077\textsuperscript{\S}  & 0.8490\textsuperscript{\ddag} &0.8737\textsuperscript{\ddag} &0.9499\textsuperscript{\S}\\\cline{2-6}
& DeepMatcher\cite{deepmatcher} (50\% training data)  &0.9347$\pm$0.0034   &0.9439$\pm$0.0074 &0.9385$\pm$0.0019 &0.9770$\pm$0.0006\\&
 Ditto \cite{ditto} (50\% training data)  &0.9356$\pm$0.0030   &\textbf{0.9448$\pm$0.0065} &0.9385$\pm$0.0016 &0.9771$\pm$0.0005\\&
 DAME (50\% training data)  &\textbf{0.9367$\pm$0.0019}   &0.9411$\pm$0.0037 &\textbf{0.9389$\pm$0.0028} &\textbf{0.9771$\pm$0.0010}\\\cline{2-6}&
 DeepMatcher\cite{deepmatcher} (full training data)  &\textbf{0.9489$\pm$0.0014}   &0.9373$\pm$0.0018 &0.9431$\pm$0.0016 &0.9789$\pm$0.0006\\&
 Ditto \cite{ditto} (full training data)  &0.9358$\pm$0.0025   &\textbf{0.9542$\pm$0.0009} &0.9449$\pm$0.0008 &0.9793$\pm$0.0003\\&
 DAME (full training data) &0.9392$\pm$0.0023   &0.9537$\pm$0.0032 &\textbf{0.9464$\pm$0.0003} &\textbf{0.9798}\\ [1ex]
\hline
\multirow{7}{*}{DBLP-ACM} &DAME (ZSL)  &0.8661\textsuperscript{\dag}  & 0.9854\textsuperscript{\S}  &0.9219\textsuperscript{\ddag} &0.9651\textsuperscript{\S}\\\cline{2-6}
& DeepMatcher\cite{deepmatcher} (50\% training data)  &0.9787$\pm$0.0098   &0.9763$\pm$0.0056 &0.9774$\pm$0.0020 &0.9919$\pm$0.0008\\&
 Ditto \cite{ditto} (50\% training data)  &\textbf{0.9865$\pm$0.0066}   &\textbf{0.9865$\pm$0.0011} &\textbf{0.9865$\pm$0.0027} &\textbf{0.9951$\pm$0.0010}\\&
  DAME (50\% training data)  &0.9787$\pm$0.0055   &0.9831$\pm$0.0011 &0.9809$\pm$0.0033 &0.9931$\pm$0.0012\\\cline{2-6}&
 DeepMatcher\cite{deepmatcher} (full training data)  & 0.9855$\pm$0.0056  & 0.9869$\pm$0.0022 &0.9861$\pm$0.0039 &0.9945$\pm$0.0014\\&
 Ditto \cite{ditto} (full training data)  & \textbf{0.9865$\pm$0.0011}  &0.9865$\pm$0.0022 &0.9865$\pm$0.0016 &0.9951$\pm$0.0006\\&
 DAME (full training data) &0.9865$\pm$0.0032   &\textbf{0.9868$\pm$0.0033} &\textbf{0.9866} &\textbf{0.9951}\\ [1ex]
\hline
\end{tabular}
\label{wild_results}
\end{table*}
\clearpage


\bibliographystyle{ACM-Reference-Format}
\bibliography{ref}


\begin{thebibliography}{53}


\ifx \showCODEN    \undefined \def \showCODEN     #1{\unskip}     \fi
\ifx \showDOI      \undefined \def \showDOI       #1{#1}\fi
\ifx \showISBNx    \undefined \def \showISBNx     #1{\unskip}     \fi
\ifx \showISBNxiii \undefined \def \showISBNxiii  #1{\unskip}     \fi
\ifx \showISSN     \undefined \def \showISSN      #1{\unskip}     \fi
\ifx \showLCCN     \undefined \def \showLCCN      #1{\unskip}     \fi
\ifx \shownote     \undefined \def \shownote      #1{#1}          \fi
\ifx \showarticletitle \undefined \def \showarticletitle #1{#1}   \fi
\ifx \showURL      \undefined \def \showURL       {\relax}        \fi
\providecommand\bibfield[2]{#2}
\providecommand\bibinfo[2]{#2}
\providecommand\natexlab[1]{#1}
\providecommand\showeprint[2][]{arXiv:#2}

\bibitem[\protect\citeauthoryear{Barlaug and Gulla}{Barlaug and Gulla}{2021}]%
        {em_survey}
\bibfield{author}{\bibinfo{person}{Nils Barlaug} {and}
  \bibinfo{person}{Jon~Atle Gulla}.} \bibinfo{year}{2021}\natexlab{}.
\newblock \showarticletitle{Neural Networks for Entity Matching: {A} Survey}.
\newblock \bibinfo{journal}{\emph{{ACM} Trans. Knowl. Discov. Data}}
  \bibinfo{volume}{15}, \bibinfo{number}{3} (\bibinfo{year}{2021}),
  \bibinfo{pages}{52:1--52:37}.
\newblock


\bibitem[\protect\citeauthoryear{Ben{-}David, Blitzer, Crammer, and
  Pereira}{Ben{-}David et~al\mbox{.}}{2006}]%
        {da1}
\bibfield{author}{\bibinfo{person}{Shai Ben{-}David}, \bibinfo{person}{John
  Blitzer}, \bibinfo{person}{Koby Crammer}, {and} \bibinfo{person}{Fernando
  Pereira}.} \bibinfo{year}{2006}\natexlab{}.
\newblock \showarticletitle{Analysis of Representations for Domain Adaptation}.
  In \bibinfo{booktitle}{\emph{Advances in Neural Information Processing
  Systems 19}}, \bibfield{editor}{\bibinfo{person}{Bernhard Sch{\"{o}}lkopf},
  \bibinfo{person}{John~C. Platt}, {and} \bibinfo{person}{Thomas Hofmann}}
  (Eds.). \bibinfo{publisher}{{MIT} Press}, \bibinfo{pages}{137--144}.
\newblock


\bibitem[\protect\citeauthoryear{Bengio, Louradour, Collobert, and
  Weston}{Bengio et~al\mbox{.}}{2009}]%
        {curriculum_learning}
\bibfield{author}{\bibinfo{person}{Yoshua Bengio},
  \bibinfo{person}{J{\'{e}}r{\^{o}}me Louradour}, \bibinfo{person}{Ronan
  Collobert}, {and} \bibinfo{person}{Jason Weston}.}
  \bibinfo{year}{2009}\natexlab{}.
\newblock \showarticletitle{Curriculum learning}. In
  \bibinfo{booktitle}{\emph{Proceedings of the 26th Annual International
  Conference on Machine Learning, {ICML}}}, Vol.~\bibinfo{volume}{382}.
  \bibinfo{publisher}{{ACM}}, \bibinfo{pages}{41--48}.
\newblock


\bibitem[\protect\citeauthoryear{Bilenko and Mooney}{Bilenko and
  Mooney}{2003}]%
        {classifier2}
\bibfield{author}{\bibinfo{person}{Mikhail Bilenko} {and}
  \bibinfo{person}{Raymond~J. Mooney}.} \bibinfo{year}{2003}\natexlab{}.
\newblock \showarticletitle{Adaptive duplicate detection using learnable string
  similarity measures}. In \bibinfo{booktitle}{\emph{Proceedings of the Ninth
  {SIGKDD} International Conference on Knowledge Discovery and Data Mining}}.
  \bibinfo{publisher}{{ACM}}, \bibinfo{pages}{39--48}.
\newblock


\bibitem[\protect\citeauthoryear{Bojanowski, Grave, Joulin, and
  Mikolov}{Bojanowski et~al\mbox{.}}{2017}]%
        {fasttext}
\bibfield{author}{\bibinfo{person}{Piotr Bojanowski}, \bibinfo{person}{Edouard
  Grave}, \bibinfo{person}{Armand Joulin}, {and} \bibinfo{person}{Tom{\'{a}}s
  Mikolov}.} \bibinfo{year}{2017}\natexlab{}.
\newblock \showarticletitle{Enriching Word Vectors with Subword Information}.
\newblock \bibinfo{journal}{\emph{Trans. Assoc. Comput. Linguistics}}
  \bibinfo{volume}{5} (\bibinfo{year}{2017}), \bibinfo{pages}{135--146}.
\newblock


\bibitem[\protect\citeauthoryear{Chen, Xu, Weinberger, and Sha}{Chen
  et~al\mbox{.}}{2012}]%
        {autoencoder}
\bibfield{author}{\bibinfo{person}{Minmin Chen},
  \bibinfo{person}{Zhixiang~Eddie Xu}, \bibinfo{person}{Kilian~Q. Weinberger},
  {and} \bibinfo{person}{Fei Sha}.} \bibinfo{year}{2012}\natexlab{}.
\newblock \showarticletitle{Marginalized Denoising Autoencoders for Domain
  Adaptation}. In \bibinfo{booktitle}{\emph{Proceedings of the 29th
  International Conference on Machine Learning, {ICML} 2012}}.
  \bibinfo{publisher}{icml.cc / Omnipress}.
\newblock


\bibitem[\protect\citeauthoryear{Chen, Trabelsi, Heflin, Xu, and Davison}{Chen
  et~al\mbox{.}}{2020}]%
        {Chen2020TableSU}
\bibfield{author}{\bibinfo{person}{Zhiyu Chen}, \bibinfo{person}{Mohamed
  Trabelsi}, \bibinfo{person}{Jeff Heflin}, \bibinfo{person}{Yinan Xu}, {and}
  \bibinfo{person}{Brian~D. Davison}.} \bibinfo{year}{2020}\natexlab{}.
\newblock \showarticletitle{Table Search Using a Deep Contextualized Language
  Model}. In \bibinfo{booktitle}{\emph{Proceedings of the 43rd International
  ACM SIGIR Conference on Research and Development in Information Retrieval}}.
  \bibinfo{publisher}{Association for Computing Machinery},
  \bibinfo{pages}{589–598}.
\newblock


\bibitem[\protect\citeauthoryear{Christen}{Christen}{2008}]%
        {classifier1}
\bibfield{author}{\bibinfo{person}{Peter Christen}.}
  \bibinfo{year}{2008}\natexlab{}.
\newblock \showarticletitle{Febrl -: an open source data cleaning,
  deduplication and record linkage system with a graphical user interface}. In
  \bibinfo{booktitle}{\emph{Proceedings of the 14th {ACM} {SIGKDD}
  International Conference on Knowledge Discovery and Data Mining}}.
  \bibinfo{publisher}{{ACM}}, \bibinfo{pages}{1065--1068}.
\newblock


\bibitem[\protect\citeauthoryear{Christen}{Christen}{2012a}]%
        {string_similarity1}
\bibfield{author}{\bibinfo{person}{Peter Christen}.}
  \bibinfo{year}{2012}\natexlab{a}.
\newblock \bibinfo{booktitle}{\emph{Data Matching - Concepts and Techniques for
  Record Linkage, Entity Resolution, and Duplicate Detection}}.
\newblock \bibinfo{publisher}{Springer}.
\newblock


\bibitem[\protect\citeauthoryear{Christen}{Christen}{2012b}]%
        {blocking1}
\bibfield{author}{\bibinfo{person}{Peter Christen}.}
  \bibinfo{year}{2012}\natexlab{b}.
\newblock \showarticletitle{A Survey of Indexing Techniques for Scalable Record
  Linkage and Deduplication}.
\newblock \bibinfo{journal}{\emph{{IEEE} Trans. Knowl. Data Eng.}}
  \bibinfo{volume}{24}, \bibinfo{number}{9} (\bibinfo{year}{2012}),
  \bibinfo{pages}{1537--1555}.
\newblock


\bibitem[\protect\citeauthoryear{Dalvi, Rastogi, Dasgupta, Sarma, and
  Sarl{\'{o}}s}{Dalvi et~al\mbox{.}}{2013}]%
        {rule1}
\bibfield{author}{\bibinfo{person}{Nilesh~N. Dalvi}, \bibinfo{person}{Vibhor
  Rastogi}, \bibinfo{person}{Anirban Dasgupta}, \bibinfo{person}{Anish~Das
  Sarma}, {and} \bibinfo{person}{Tam{\'{a}}s Sarl{\'{o}}s}.}
  \bibinfo{year}{2013}\natexlab{}.
\newblock \showarticletitle{Optimal hashing schemes for entity matching}. In
  \bibinfo{booktitle}{\emph{22nd International World Wide Web Conference,
  {WWW}}}. \bibinfo{pages}{295--306}.
\newblock


\bibitem[\protect\citeauthoryear{Devlin, Chang, Lee, and Toutanova}{Devlin
  et~al\mbox{.}}{2019}]%
        {Devlin2019BERTPO}
\bibfield{author}{\bibinfo{person}{Jacob Devlin}, \bibinfo{person}{Ming-Wei
  Chang}, \bibinfo{person}{Kenton Lee}, {and} \bibinfo{person}{Kristina
  Toutanova}.} \bibinfo{year}{2019}\natexlab{}.
\newblock \showarticletitle{BERT: Pre-training of Deep Bidirectional
  Transformers for Language Understanding}. In
  \bibinfo{booktitle}{\emph{NAACL-HLT}}.
\newblock


\bibitem[\protect\citeauthoryear{Ebraheem, Thirumuruganathan, Joty, Ouzzani,
  and Tang}{Ebraheem et~al\mbox{.}}{2018}]%
        {ebraheem_vldb}
\bibfield{author}{\bibinfo{person}{Muhammad Ebraheem},
  \bibinfo{person}{Saravanan Thirumuruganathan}, \bibinfo{person}{Shafiq~R.
  Joty}, \bibinfo{person}{Mourad Ouzzani}, {and} \bibinfo{person}{Nan Tang}.}
  \bibinfo{year}{2018}\natexlab{}.
\newblock \showarticletitle{Distributed Representations of Tuples for Entity
  Resolution}.
\newblock \bibinfo{journal}{\emph{Proc. {VLDB} Endow.}} \bibinfo{volume}{11},
  \bibinfo{number}{11} (\bibinfo{year}{2018}), \bibinfo{pages}{1454--1467}.
\newblock


\bibitem[\protect\citeauthoryear{Elmagarmid, Ilyas, Ouzzani,
  Quian{\'{e}}{-}Ruiz, Tang, and Yin}{Elmagarmid et~al\mbox{.}}{2014}]%
        {rule2}
\bibfield{author}{\bibinfo{person}{Ahmed~K. Elmagarmid},
  \bibinfo{person}{Ihab~F. Ilyas}, \bibinfo{person}{Mourad Ouzzani},
  \bibinfo{person}{Jorge{-}Arnulfo Quian{\'{e}}{-}Ruiz}, \bibinfo{person}{Nan
  Tang}, {and} \bibinfo{person}{Si Yin}.} \bibinfo{year}{2014}\natexlab{}.
\newblock \showarticletitle{{NADEEF/ER:} generic and interactive entity
  resolution}. In \bibinfo{booktitle}{\emph{International Conference on
  Management of Data, {SIGMOD} 2014}}. \bibinfo{publisher}{{ACM}},
  \bibinfo{pages}{1071--1074}.
\newblock


\bibitem[\protect\citeauthoryear{Elmagarmid, Ipeirotis, and
  Verykios}{Elmagarmid et~al\mbox{.}}{2007}]%
        {string_similarity2}
\bibfield{author}{\bibinfo{person}{Ahmed~K. Elmagarmid},
  \bibinfo{person}{Panagiotis~G. Ipeirotis}, {and}
  \bibinfo{person}{Vassilios~S. Verykios}.} \bibinfo{year}{2007}\natexlab{}.
\newblock \showarticletitle{Duplicate Record Detection: {A} Survey}.
\newblock \bibinfo{journal}{\emph{{IEEE} Trans. Knowl. Data Eng.}}
  \bibinfo{volume}{19}, \bibinfo{number}{1} (\bibinfo{year}{2007}),
  \bibinfo{pages}{1--16}.
\newblock


\bibitem[\protect\citeauthoryear{Fisher, Christen, Wang, and Rahm}{Fisher
  et~al\mbox{.}}{2015}]%
        {blocking2}
\bibfield{author}{\bibinfo{person}{Jeffrey Fisher}, \bibinfo{person}{Peter
  Christen}, \bibinfo{person}{Qing Wang}, {and} \bibinfo{person}{Erhard Rahm}.}
  \bibinfo{year}{2015}\natexlab{}.
\newblock \showarticletitle{A Clustering-Based Framework to Control Block Sizes
  for Entity Resolution}. In \bibinfo{booktitle}{\emph{Proceedings of the 21th
  {ACM} {SIGKDD} International Conference on Knowledge Discovery and Data
  Mining, Sydney, NSW, Australia, August 10-13, 2015}}.
  \bibinfo{publisher}{{ACM}}, \bibinfo{pages}{279--288}.
\newblock


\bibitem[\protect\citeauthoryear{Fu, Han, Sun, Chen, Zhang, Wu, and Kong}{Fu
  et~al\mbox{.}}{2019}]%
        {mpm}
\bibfield{author}{\bibinfo{person}{Cheng Fu}, \bibinfo{person}{Xianpei Han},
  \bibinfo{person}{Le Sun}, \bibinfo{person}{Bo Chen}, \bibinfo{person}{Wei
  Zhang}, \bibinfo{person}{Suhui Wu}, {and} \bibinfo{person}{Hao Kong}.}
  \bibinfo{year}{2019}\natexlab{}.
\newblock \showarticletitle{End-to-End Multi-Perspective Matching for Entity
  Resolution}. In \bibinfo{booktitle}{\emph{Proceedings of the Twenty-Eighth
  International Joint Conference on Artificial Intelligence, {IJCAI} 2019}}.
  \bibinfo{publisher}{ijcai.org}, \bibinfo{pages}{4961--4967}.
\newblock


\bibitem[\protect\citeauthoryear{Gal, Islam, and Ghahramani}{Gal
  et~al\mbox{.}}{2017}]%
        {usde_bald}
\bibfield{author}{\bibinfo{person}{Yarin Gal}, \bibinfo{person}{Riashat Islam},
  {and} \bibinfo{person}{Zoubin Ghahramani}.} \bibinfo{year}{2017}\natexlab{}.
\newblock \showarticletitle{Deep Bayesian Active Learning with Image Data}. In
  \bibinfo{booktitle}{\emph{Proceedings of the 34th International Conference on
  Machine Learning, {ICML} 2017, Sydney, NSW, Australia, 6-11 August 2017}}
  \emph{(\bibinfo{series}{Proceedings of Machine Learning Research},
  Vol.~\bibinfo{volume}{70})}. \bibinfo{publisher}{{PMLR}},
  \bibinfo{pages}{1183--1192}.
\newblock


\bibitem[\protect\citeauthoryear{Guo, Shah, and Barzilay}{Guo
  et~al\mbox{.}}{2018}]%
        {guo_emnlp}
\bibfield{author}{\bibinfo{person}{Jiang Guo}, \bibinfo{person}{Darsh~J. Shah},
  {and} \bibinfo{person}{Regina Barzilay}.} \bibinfo{year}{2018}\natexlab{}.
\newblock \showarticletitle{Multi-Source Domain Adaptation with Mixture of
  Experts}. In \bibinfo{booktitle}{\emph{Proceedings of the 2018 Conference on
  Empirical Methods in Natural Language Processing}}.
  \bibinfo{publisher}{Association for Computational Linguistics},
  \bibinfo{pages}{4694--4703}.
\newblock


\bibitem[\protect\citeauthoryear{Gururangan, Marasovic, Swayamdipta, Lo,
  Beltagy, Downey, and Smith}{Gururangan et~al\mbox{.}}{2020}]%
        {gururangan_acl}
\bibfield{author}{\bibinfo{person}{Suchin Gururangan}, \bibinfo{person}{Ana
  Marasovic}, \bibinfo{person}{Swabha Swayamdipta}, \bibinfo{person}{Kyle Lo},
  \bibinfo{person}{Iz Beltagy}, \bibinfo{person}{Doug Downey}, {and}
  \bibinfo{person}{Noah~A. Smith}.} \bibinfo{year}{2020}\natexlab{}.
\newblock \showarticletitle{Don't Stop Pretraining: Adapt Language Models to
  Domains and Tasks}. In \bibinfo{booktitle}{\emph{Proceedings of the 58th
  Annual Meeting of the Association for Computational Linguistics, {ACL}
  2020}}. \bibinfo{publisher}{Association for Computational Linguistics},
  \bibinfo{pages}{8342--8360}.
\newblock


\bibitem[\protect\citeauthoryear{Han and Eisenstein}{Han and
  Eisenstein}{2019}]%
        {han_emnlp}
\bibfield{author}{\bibinfo{person}{Xiaochuang Han} {and} \bibinfo{person}{Jacob
  Eisenstein}.} \bibinfo{year}{2019}\natexlab{}.
\newblock \showarticletitle{Unsupervised Domain Adaptation of Contextualized
  Embeddings for Sequence Labeling}. In \bibinfo{booktitle}{\emph{Proceedings
  of the 2019 Conference on Empirical Methods in Natural Language Processing
  and the 9th International Joint Conference on Natural Language Processing,
  {EMNLP-IJCNLP} 2019}}. \bibinfo{publisher}{Association for Computational
  Linguistics}, \bibinfo{pages}{4237--4247}.
\newblock


\bibitem[\protect\citeauthoryear{Kasai, Qian, Gurajada, Li, and Popa}{Kasai
  et~al\mbox{.}}{2019}]%
        {active_learning_em}
\bibfield{author}{\bibinfo{person}{Jungo Kasai}, \bibinfo{person}{Kun Qian},
  \bibinfo{person}{Sairam Gurajada}, \bibinfo{person}{Yunyao Li}, {and}
  \bibinfo{person}{Lucian Popa}.} \bibinfo{year}{2019}\natexlab{}.
\newblock \showarticletitle{Low-resource Deep Entity Resolution with Transfer
  and Active Learning}. In \bibinfo{booktitle}{\emph{Proceedings of the 57th
  Conference of the Association for Computational Linguistics, {ACL} 2019,
  Florence, Italy, July 28- August 2, 2019, Volume 1: Long Papers}}.
  \bibinfo{publisher}{Association for Computational Linguistics},
  \bibinfo{pages}{5851--5861}.
\newblock


\bibitem[\protect\citeauthoryear{Kim, Stratos, and Kim}{Kim
  et~al\mbox{.}}{2017}]%
        {expert1}
\bibfield{author}{\bibinfo{person}{Young{-}Bum Kim}, \bibinfo{person}{Karl
  Stratos}, {and} \bibinfo{person}{Dongchan Kim}.}
  \bibinfo{year}{2017}\natexlab{}.
\newblock \showarticletitle{Domain Attention with an Ensemble of Experts}. In
  \bibinfo{booktitle}{\emph{Proceedings of the 55th Annual Meeting of the
  Association for Computational Linguistics, {ACL} 2017}}.
  \bibinfo{pages}{643--653}.
\newblock


\bibitem[\protect\citeauthoryear{Konda, Das, C., Doan, Ardalan, Ballard, Li,
  Panahi, Zhang, Naughton, Prasad, Krishnan, Deep, and Raghavendra}{Konda
  et~al\mbox{.}}{2016}]%
        {magellan}
\bibfield{author}{\bibinfo{person}{Pradap Konda}, \bibinfo{person}{Sanjib Das},
  \bibinfo{person}{Paul Suganthan~G. C.}, \bibinfo{person}{AnHai Doan},
  \bibinfo{person}{Adel Ardalan}, \bibinfo{person}{Jeffrey~R. Ballard},
  \bibinfo{person}{Han Li}, \bibinfo{person}{Fatemah Panahi},
  \bibinfo{person}{Haojun Zhang}, \bibinfo{person}{Jeffrey~F. Naughton},
  \bibinfo{person}{Shishir Prasad}, \bibinfo{person}{Ganesh Krishnan},
  \bibinfo{person}{Rohit Deep}, {and} \bibinfo{person}{Vijay Raghavendra}.}
  \bibinfo{year}{2016}\natexlab{}.
\newblock \showarticletitle{Magellan: Toward Building Entity Matching
  Management Systems}.
\newblock \bibinfo{journal}{\emph{Proc. {VLDB} Endow.}} \bibinfo{volume}{9},
  \bibinfo{number}{12} (\bibinfo{year}{2016}), \bibinfo{pages}{1197--1208}.
\newblock


\bibitem[\protect\citeauthoryear{K{\"{o}}pcke, Thor, and Rahm}{K{\"{o}}pcke
  et~al\mbox{.}}{2010}]%
        {er_datasets}
\bibfield{author}{\bibinfo{person}{Hanna K{\"{o}}pcke},
  \bibinfo{person}{Andreas Thor}, {and} \bibinfo{person}{Erhard Rahm}.}
  \bibinfo{year}{2010}\natexlab{}.
\newblock \showarticletitle{Evaluation of entity resolution approaches on
  real-world match problems}.
\newblock \bibinfo{journal}{\emph{Proc. {VLDB} Endow.}} \bibinfo{volume}{3},
  \bibinfo{number}{1} (\bibinfo{year}{2010}), \bibinfo{pages}{484--493}.
\newblock


\bibitem[\protect\citeauthoryear{Li, Li, Suhara, Doan, and Tan}{Li
  et~al\mbox{.}}{2020}]%
        {ditto}
\bibfield{author}{\bibinfo{person}{Yuliang Li}, \bibinfo{person}{Jinfeng Li},
  \bibinfo{person}{Yoshihiko Suhara}, \bibinfo{person}{AnHai Doan}, {and}
  \bibinfo{person}{Wang{-}Chiew Tan}.} \bibinfo{year}{2020}\natexlab{}.
\newblock \showarticletitle{Deep Entity Matching with Pre-Trained Language
  Models}.
\newblock \bibinfo{journal}{\emph{Proc. {VLDB} Endow.}} \bibinfo{volume}{14},
  \bibinfo{number}{1} (\bibinfo{year}{2020}), \bibinfo{pages}{50--60}.
\newblock
\urldef\tempurl%
\url{https://doi.org/10.14778/3421424.3421431}
\showDOI{\tempurl}


\bibitem[\protect\citeauthoryear{Liu, Ott, Goyal, Du, Joshi, Chen, Levy, Lewis,
  Zettlemoyer, and Stoyanov}{Liu et~al\mbox{.}}{2019}]%
        {Liu2019RoBERTaAR}
\bibfield{author}{\bibinfo{person}{Yinhan Liu}, \bibinfo{person}{Myle Ott},
  \bibinfo{person}{Naman Goyal}, \bibinfo{person}{Jingfei Du},
  \bibinfo{person}{Mandar Joshi}, \bibinfo{person}{Danqi Chen},
  \bibinfo{person}{Omer Levy}, \bibinfo{person}{Mike Lewis},
  \bibinfo{person}{Luke Zettlemoyer}, {and} \bibinfo{person}{Veselin
  Stoyanov}.} \bibinfo{year}{2019}\natexlab{}.
\newblock \showarticletitle{RoBERTa: A Robustly Optimized BERT Pretraining
  Approach}.
\newblock \bibinfo{journal}{\emph{ArXiv}}  \bibinfo{volume}{abs/1907.11692}
  (\bibinfo{year}{2019}).
\newblock


\bibitem[\protect\citeauthoryear{Lu, Lin, Wang, and Li}{Lu
  et~al\mbox{.}}{2019}]%
        {string_similarity3}
\bibfield{author}{\bibinfo{person}{Jiaheng Lu}, \bibinfo{person}{Chunbin Lin},
  \bibinfo{person}{Jin Wang}, {and} \bibinfo{person}{Chen Li}.}
  \bibinfo{year}{2019}\natexlab{}.
\newblock \showarticletitle{Synergy of Database Techniques and Machine Learning
  Models for String Similarity Search and Join}. In
  \bibinfo{booktitle}{\emph{Proceedings of the 28th {ACM} International
  Conference on Information and Knowledge Management, {CIKM} 2019}}.
  \bibinfo{publisher}{{ACM}}, \bibinfo{pages}{2975--2976}.
\newblock


\bibitem[\protect\citeauthoryear{Ma, Xu, Wang, Nallapati, and Xiang}{Ma
  et~al\mbox{.}}{2019}]%
        {ma_emnlp}
\bibfield{author}{\bibinfo{person}{Xiaofei Ma}, \bibinfo{person}{Peng Xu},
  \bibinfo{person}{Zhiguo Wang}, \bibinfo{person}{Ramesh Nallapati}, {and}
  \bibinfo{person}{Bing Xiang}.} \bibinfo{year}{2019}\natexlab{}.
\newblock \showarticletitle{Domain Adaptation with BERT-based Domain
  Classification and Data Selection}. In \bibinfo{booktitle}{\emph{Proceedings
  of the 2nd Workshop on Deep Learning Approaches for Low-Resource NLP}}.
  \bibinfo{publisher}{Association for Computational Linguistics},
  \bibinfo{pages}{76--83}.
\newblock


\bibitem[\protect\citeauthoryear{Mudgal, Li, Rekatsinas, Doan, Park, Krishnan,
  Deep, Arcaute, and Raghavendra}{Mudgal et~al\mbox{.}}{2018}]%
        {deepmatcher}
\bibfield{author}{\bibinfo{person}{Sidharth Mudgal}, \bibinfo{person}{Han Li},
  \bibinfo{person}{Theodoros Rekatsinas}, \bibinfo{person}{AnHai Doan},
  \bibinfo{person}{Youngchoon Park}, \bibinfo{person}{Ganesh Krishnan},
  \bibinfo{person}{Rohit Deep}, \bibinfo{person}{Esteban Arcaute}, {and}
  \bibinfo{person}{Vijay Raghavendra}.} \bibinfo{year}{2018}\natexlab{}.
\newblock \showarticletitle{Deep Learning for Entity Matching: {A} Design Space
  Exploration}. In \bibinfo{booktitle}{\emph{Proceedings of the 2018
  International Conference on Management of Data, {SIGMOD} Conference 2018,
  Houston, TX, USA, June 10-15, 2018}}. \bibinfo{publisher}{{ACM}},
  \bibinfo{pages}{19--34}.
\newblock


\bibitem[\protect\citeauthoryear{Nickel, Murphy, Tresp, and Gabrilovich}{Nickel
  et~al\mbox{.}}{2016}]%
        {kb_enrich}
\bibfield{author}{\bibinfo{person}{Maximilian Nickel}, \bibinfo{person}{Kevin
  Murphy}, \bibinfo{person}{Volker Tresp}, {and} \bibinfo{person}{Evgeniy
  Gabrilovich}.} \bibinfo{year}{2016}\natexlab{}.
\newblock \showarticletitle{A Review of Relational Machine Learning for
  Knowledge Graphs}.
\newblock \bibinfo{journal}{\emph{Proc. {IEEE}}} \bibinfo{volume}{104},
  \bibinfo{number}{1} (\bibinfo{year}{2016}), \bibinfo{pages}{11--33}.
\newblock


\bibitem[\protect\citeauthoryear{Pan, Ni, Sun, Yang, and Chen}{Pan
  et~al\mbox{.}}{2010}]%
        {da2}
\bibfield{author}{\bibinfo{person}{Sinno~Jialin Pan},
  \bibinfo{person}{Xiaochuan Ni}, \bibinfo{person}{Jian{-}Tao Sun},
  \bibinfo{person}{Qiang Yang}, {and} \bibinfo{person}{Zheng Chen}.}
  \bibinfo{year}{2010}\natexlab{}.
\newblock \showarticletitle{Cross-domain sentiment classification via spectral
  feature alignment}. In \bibinfo{booktitle}{\emph{Proceedings of the 19th
  International Conference on World Wide Web, {WWW} 2010}}.
  \bibinfo{publisher}{{ACM}}, \bibinfo{pages}{751--760}.
\newblock


\bibitem[\protect\citeauthoryear{Papadakis, Skoutas, Thanos, and
  Palpanas}{Papadakis et~al\mbox{.}}{2020}]%
        {blocking3}
\bibfield{author}{\bibinfo{person}{George Papadakis},
  \bibinfo{person}{Dimitrios Skoutas}, \bibinfo{person}{Emmanouil Thanos},
  {and} \bibinfo{person}{Themis Palpanas}.} \bibinfo{year}{2020}\natexlab{}.
\newblock \showarticletitle{Blocking and Filtering Techniques for Entity
  Resolution: {A} Survey}.
\newblock \bibinfo{journal}{\emph{{ACM} Comput. Surv.}} \bibinfo{volume}{53},
  \bibinfo{number}{2} (\bibinfo{year}{2020}), \bibinfo{pages}{31:1--31:42}.
\newblock


\bibitem[\protect\citeauthoryear{Parikh, T{\"{a}}ckstr{\"{o}}m, Das, and
  Uszkoreit}{Parikh et~al\mbox{.}}{2016}]%
        {decom_attention}
\bibfield{author}{\bibinfo{person}{Ankur~P. Parikh}, \bibinfo{person}{Oscar
  T{\"{a}}ckstr{\"{o}}m}, \bibinfo{person}{Dipanjan Das}, {and}
  \bibinfo{person}{Jakob Uszkoreit}.} \bibinfo{year}{2016}\natexlab{}.
\newblock \showarticletitle{A Decomposable Attention Model for Natural Language
  Inference}. In \bibinfo{booktitle}{\emph{Proceedings of the 2016 Conference
  on Empirical Methods in Natural Language Processing, {EMNLP} 2016, 1-4,
  2016}}. \bibinfo{publisher}{The Association for Computational Linguistics},
  \bibinfo{pages}{2249--2255}.
\newblock


\bibitem[\protect\citeauthoryear{Pennington, Socher, and Manning}{Pennington
  et~al\mbox{.}}{2014}]%
        {glove}
\bibfield{author}{\bibinfo{person}{Jeffrey Pennington},
  \bibinfo{person}{Richard Socher}, {and} \bibinfo{person}{Christopher
  Manning}.} \bibinfo{year}{2014}\natexlab{}.
\newblock \showarticletitle{{G}lo{V}e: Global Vectors for Word Representation}.
  In \bibinfo{booktitle}{\emph{Proceedings of the 2014 Conference on Empirical
  Methods in Natural Language Processing ({EMNLP})}}.
  \bibinfo{publisher}{Association for Computational Linguistics},
  \bibinfo{pages}{1532--1543}.
\newblock


\bibitem[\protect\citeauthoryear{Rietzler, Stabinger, Opitz, and Engl}{Rietzler
  et~al\mbox{.}}{2020}]%
        {rietzler_lrec}
\bibfield{author}{\bibinfo{person}{Alexander Rietzler},
  \bibinfo{person}{Sebastian Stabinger}, \bibinfo{person}{Paul Opitz}, {and}
  \bibinfo{person}{Stefan Engl}.} \bibinfo{year}{2020}\natexlab{}.
\newblock \showarticletitle{Adapt or Get Left Behind: Domain Adaptation through
  {BERT} Language Model Finetuning for Aspect-Target Sentiment Classification}.
  In \bibinfo{booktitle}{\emph{Proceedings of The 12th Language Resources and
  Evaluation Conference, {LREC} 2020}}. \bibinfo{publisher}{European Language
  Resources Association}, \bibinfo{pages}{4933--4941}.
\newblock


\bibitem[\protect\citeauthoryear{Sakata, Shibata, Tanaka, and Kurohashi}{Sakata
  et~al\mbox{.}}{2019}]%
        {sakata2019}
\bibfield{author}{\bibinfo{person}{Wataru Sakata}, \bibinfo{person}{Tomohide
  Shibata}, \bibinfo{person}{Ribeka Tanaka}, {and} \bibinfo{person}{Sadao
  Kurohashi}.} \bibinfo{year}{2019}\natexlab{}.
\newblock \showarticletitle{FAQ Retrieval Using Query-Question Similarity and
  BERT-Based Query-Answer Relevance}. In \bibinfo{booktitle}{\emph{Proceedings
  of the 42nd International ACM SIGIR Conference on Research and Development in
  Information Retrieval}}. \bibinfo{pages}{1113–1116}.
\newblock


\bibitem[\protect\citeauthoryear{Sanh, Debut, Chaumond, and Wolf}{Sanh
  et~al\mbox{.}}{2019}]%
        {distilbert}
\bibfield{author}{\bibinfo{person}{Victor Sanh}, \bibinfo{person}{Lysandre
  Debut}, \bibinfo{person}{Julien Chaumond}, {and} \bibinfo{person}{Thomas
  Wolf}.} \bibinfo{year}{2019}\natexlab{}.
\newblock \showarticletitle{DistilBERT, a distilled version of {BERT:} smaller,
  faster, cheaper and lighter}.
\newblock \bibinfo{journal}{\emph{CoRR}}  \bibinfo{volume}{abs/1910.01108}
  (\bibinfo{year}{2019}).
\newblock


\bibitem[\protect\citeauthoryear{Sener and Savarese}{Sener and
  Savarese}{2018}]%
        {coreset}
\bibfield{author}{\bibinfo{person}{Ozan Sener} {and} \bibinfo{person}{Silvio
  Savarese}.} \bibinfo{year}{2018}\natexlab{}.
\newblock \showarticletitle{Active Learning for Convolutional Neural Networks:
  {A} Core-Set Approach}. In \bibinfo{booktitle}{\emph{6th International
  Conference on Learning Representations, {ICLR} 2018, Vancouver, BC, Canada,
  April 30 - May 3, 2018, Conference Track Proceedings}}.
  \bibinfo{publisher}{OpenReview.net}.
\newblock


\bibitem[\protect\citeauthoryear{Shen, Qu, Zhang, and Yu}{Shen
  et~al\mbox{.}}{[n.d.]}]%
        {adv2}
\bibfield{author}{\bibinfo{person}{Jian Shen}, \bibinfo{person}{Yanru Qu},
  \bibinfo{person}{Weinan Zhang}, {and} \bibinfo{person}{Yong Yu}.}
  \bibinfo{year}{[n.d.]}\natexlab{}.
\newblock \showarticletitle{Wasserstein Distance Guided Representation Learning
  for Domain Adaptation}. In \bibinfo{booktitle}{\emph{Proceedings of the
  Thirty-Second {AAAI} Conference on Artificial Intelligence, (AAAI-18), the
  30th innovative Applications of Artificial Intelligence (IAAI-18)}}.
  \bibinfo{pages}{4058--4065}.
\newblock


\bibitem[\protect\citeauthoryear{Shen, Wang, and Han}{Shen
  et~al\mbox{.}}{2015}]%
        {entity_linking}
\bibfield{author}{\bibinfo{person}{Wei Shen}, \bibinfo{person}{Jianyong Wang},
  {and} \bibinfo{person}{Jiawei Han}.} \bibinfo{year}{2015}\natexlab{}.
\newblock \showarticletitle{Entity Linking with a Knowledge Base: Issues,
  Techniques, and Solutions}.
\newblock \bibinfo{journal}{\emph{{IEEE} Trans. Knowl. Data Eng.}}
  \bibinfo{volume}{27}, \bibinfo{number}{2} (\bibinfo{year}{2015}),
  \bibinfo{pages}{443--460}.
\newblock


\bibitem[\protect\citeauthoryear{Singh, Meduri, Elmagarmid, Madden, Papotti,
  Quian{\'{e}}{-}Ruiz, Solar{-}Lezama, and Tang}{Singh et~al\mbox{.}}{2017}]%
        {rule3}
\bibfield{author}{\bibinfo{person}{Rohit Singh},
  \bibinfo{person}{Venkata~Vamsikrishna Meduri}, \bibinfo{person}{Ahmed~K.
  Elmagarmid}, \bibinfo{person}{Samuel Madden}, \bibinfo{person}{Paolo
  Papotti}, \bibinfo{person}{Jorge{-}Arnulfo Quian{\'{e}}{-}Ruiz},
  \bibinfo{person}{Armando Solar{-}Lezama}, {and} \bibinfo{person}{Nan Tang}.}
  \bibinfo{year}{2017}\natexlab{}.
\newblock \showarticletitle{Synthesizing Entity Matching Rules by Examples}.
\newblock \bibinfo{journal}{\emph{Proc. {VLDB} Endow.}} \bibinfo{volume}{11},
  \bibinfo{number}{2} (\bibinfo{year}{2017}), \bibinfo{pages}{189--202}.
\newblock


\bibitem[\protect\citeauthoryear{Sun, Feng, and Saenko}{Sun
  et~al\mbox{.}}{2016}]%
        {covariance}
\bibfield{author}{\bibinfo{person}{Baochen Sun}, \bibinfo{person}{Jiashi Feng},
  {and} \bibinfo{person}{Kate Saenko}.} \bibinfo{year}{2016}\natexlab{}.
\newblock \showarticletitle{Return of Frustratingly Easy Domain Adaptation}. In
  \bibinfo{booktitle}{\emph{Proceedings of the Thirtieth {AAAI} Conference on
  Artificial Intelligence, February 12-17, 2016}}. \bibinfo{publisher}{{AAAI}
  Press}, \bibinfo{pages}{2058--2065}.
\newblock


\bibitem[\protect\citeauthoryear{Trabelsi, Cao, and Heflin}{Trabelsi
  et~al\mbox{.}}{2020}]%
        {selab_arxiv}
\bibfield{author}{\bibinfo{person}{Mohamed Trabelsi}, \bibinfo{person}{Jin
  Cao}, {and} \bibinfo{person}{Jeff Heflin}.} \bibinfo{year}{2020}\natexlab{}.
\newblock \showarticletitle{Semantic Labeling Using a Deep Contextualized
  Language Model}.
\newblock \bibinfo{journal}{\emph{CoRR}}  \bibinfo{volume}{abs/2010.16037}
  (\bibinfo{year}{2020}).
\newblock


\bibitem[\protect\citeauthoryear{Trabelsi, Cao, and Heflin}{Trabelsi
  et~al\mbox{.}}{2021a}]%
        {selab_ijcnn}
\bibfield{author}{\bibinfo{person}{Mohamed Trabelsi}, \bibinfo{person}{Jin
  Cao}, {and} \bibinfo{person}{Jeff Heflin}.} \bibinfo{year}{2021}\natexlab{a}.
\newblock \showarticletitle{SeLaB: Semantic Labeling with {BERT}}. In
  \bibinfo{booktitle}{\emph{International Joint Conference on Neural Networks,
  {IJCNN} 2021, Shenzhen, China, July 18-22, 2021}}.
  \bibinfo{publisher}{{IEEE}}, \bibinfo{pages}{1--8}.
\newblock


\bibitem[\protect\citeauthoryear{Trabelsi, Chen, Davison, and Heflin}{Trabelsi
  et~al\mbox{.}}{2021b}]%
        {survey_doc_retrieval}
\bibfield{author}{\bibinfo{person}{Mohamed Trabelsi}, \bibinfo{person}{Zhiyu
  Chen}, \bibinfo{person}{Brian~D. Davison}, {and} \bibinfo{person}{Jeff
  Heflin}.} \bibinfo{year}{2021}\natexlab{b}.
\newblock \showarticletitle{Neural ranking models for document retrieval}.
\newblock \bibinfo{journal}{\emph{Inf. Retr. J.}} \bibinfo{volume}{24},
  \bibinfo{number}{6} (\bibinfo{year}{2021}), \bibinfo{pages}{400--444}.
\newblock


\bibitem[\protect\citeauthoryear{Trabelsi, Chen, Zhang, Davison, and
  Heflin}{Trabelsi et~al\mbox{.}}{2022}]%
        {strubert}
\bibfield{author}{\bibinfo{person}{Mohamed Trabelsi}, \bibinfo{person}{Zhiyu
  Chen}, \bibinfo{person}{Shuo Zhang}, \bibinfo{person}{Brian~D. Davison},
  {and} \bibinfo{person}{Jeff Heflin}.} \bibinfo{year}{2022}\natexlab{}.
\newblock \showarticletitle{StruBERT: Structure-aware BERT for Table Search and
  Matching}. In \bibinfo{booktitle}{\emph{Proceddings of The ACM Web Conference
  2022}}.
\newblock


\bibitem[\protect\citeauthoryear{Vaswani, Shazeer, Parmar, Uszkoreit, Jones,
  Gomez, Kaiser, and Polosukhin}{Vaswani et~al\mbox{.}}{2017}]%
        {transformer}
\bibfield{author}{\bibinfo{person}{Ashish Vaswani}, \bibinfo{person}{Noam
  Shazeer}, \bibinfo{person}{Niki Parmar}, \bibinfo{person}{Jakob Uszkoreit},
  \bibinfo{person}{Llion Jones}, \bibinfo{person}{Aidan~N Gomez},
  \bibinfo{person}{\L~ukasz Kaiser}, {and} \bibinfo{person}{Illia Polosukhin}.}
  \bibinfo{year}{2017}\natexlab{}.
\newblock \showarticletitle{Attention is All you Need}.
\newblock In \bibinfo{booktitle}{\emph{Advances in Neural Information
  Processing Systems 30}}. \bibinfo{pages}{5998--6008}.
\newblock


\bibitem[\protect\citeauthoryear{Wang, Singh, Michael, Hill, Levy, and
  Bowman}{Wang et~al\mbox{.}}{2018}]%
        {wang-etal-2018-glue}
\bibfield{author}{\bibinfo{person}{Alex Wang}, \bibinfo{person}{Amanpreet
  Singh}, \bibinfo{person}{Julian Michael}, \bibinfo{person}{Felix Hill},
  \bibinfo{person}{Omer Levy}, {and} \bibinfo{person}{Samuel Bowman}.}
  \bibinfo{year}{2018}\natexlab{}.
\newblock \showarticletitle{{GLUE}: A Multi-Task Benchmark and Analysis
  Platform for Natural Language Understanding}. In
  \bibinfo{booktitle}{\emph{Proceedings of the 2018 {EMNLP} Workshop
  {B}lackbox{NLP}: Analyzing and Interpreting Neural Networks for {NLP}}}.
  \bibinfo{publisher}{Association for Computational Linguistics},
  \bibinfo{pages}{353--355}.
\newblock


\bibitem[\protect\citeauthoryear{Wang and Shang}{Wang and Shang}{2014}]%
        {least_entropy}
\bibfield{author}{\bibinfo{person}{Dan Wang} {and} \bibinfo{person}{Yi Shang}.}
  \bibinfo{year}{2014}\natexlab{}.
\newblock \showarticletitle{A new active labeling method for deep learning}. In
  \bibinfo{booktitle}{\emph{2014 International Joint Conference on Neural
  Networks, {IJCNN} 2014, Beijing, China, July 6-11, 2014}}.
  \bibinfo{publisher}{{IEEE}}, \bibinfo{pages}{112--119}.
\newblock


\bibitem[\protect\citeauthoryear{Wright and Augenstein}{Wright and
  Augenstein}{2020}]%
        {wright_emnlp}
\bibfield{author}{\bibinfo{person}{Dustin Wright} {and}
  \bibinfo{person}{Isabelle Augenstein}.} \bibinfo{year}{2020}\natexlab{}.
\newblock \showarticletitle{Transformer Based Multi-Source Domain Adaptation}.
  In \bibinfo{booktitle}{\emph{Proceedings of the 2020 Conference on Empirical
  Methods in Natural Language Processing, {EMNLP} 2020}}.
  \bibinfo{publisher}{Association for Computational Linguistics},
  \bibinfo{pages}{7963--7974}.
\newblock


\bibitem[\protect\citeauthoryear{Zhang, Barzilay, and Jaakkola}{Zhang
  et~al\mbox{.}}{2017}]%
        {adv1}
\bibfield{author}{\bibinfo{person}{Yuan Zhang}, \bibinfo{person}{Regina
  Barzilay}, {and} \bibinfo{person}{Tommi~S. Jaakkola}.}
  \bibinfo{year}{2017}\natexlab{}.
\newblock \showarticletitle{Aspect-augmented Adversarial Networks for Domain
  Adaptation}.
\newblock \bibinfo{journal}{\emph{Trans. Assoc. Comput. Linguistics}}
  \bibinfo{volume}{5} (\bibinfo{year}{2017}), \bibinfo{pages}{515--528}.
\newblock


\bibitem[\protect\citeauthoryear{Zhao and He}{Zhao and He}{2019}]%
        {auto-em}
\bibfield{author}{\bibinfo{person}{Chen Zhao} {and} \bibinfo{person}{Yeye He}.}
  \bibinfo{year}{2019}\natexlab{}.
\newblock \showarticletitle{Auto-EM: End-to-end Fuzzy Entity-Matching using
  Pre-trained Deep Models and Transfer Learning}. In
  \bibinfo{booktitle}{\emph{The World Wide Web Conference, {WWW} 2019}}.
  \bibinfo{publisher}{{ACM}}, \bibinfo{pages}{2413--2424}.
\newblock


\end{thebibliography}
\appendix

\end{document}